\documentclass[11pt]{article}
\usepackage[preprint]{acl}
\usepackage{times}
\usepackage{latexsym}
\usepackage[T1]{fontenc}
\usepackage[utf8]{inputenc}
\usepackage{microtype}
\usepackage{inconsolata}
\usepackage{amsmath}
\usepackage{amssymb}
\usepackage{booktabs}
\usepackage{graphicx}
\usepackage{xurl}
\usepackage{float}
\usepackage{tikz}
\usetikzlibrary{arrows.meta,positioning,calc,fit,backgrounds}
\usepackage{multirow}
\usepackage{xcolor}
\usepackage{array}
\usepackage{colortbl}
\usepackage{makecell}
\usepackage{longtable}

\definecolor{avgcol}{RGB}{232,240,247}
\definecolor{avg2col}{RGB}{231,244,233}
\newcolumntype{H}{>{\columncolor{avgcol}}c}
\newcolumntype{G}{>{\columncolor{avg2col}}c}
\definecolor{hdrcol}{RGB}{236,237,242}
\definecolor{pbarcol}{RGB}{201,221,242}
\newlength{\pbw}
\newcommand{\pbar}[2]{\rlap{\textcolor{pbarcol}{\rule[-0.08ex]{#1\pbw}{1.18ex}}}\makebox[\pbw][r]{#2}}
\newcommand{\lc}[1]{\makebox[10mm][r]{#1}}
\newcommand{\uc}[1]{\makebox[18mm][r]{#1}}
\newcommand{\hc}[1]{\makebox[15mm][r]{#1}}

\newcommand{\best}[1]{\textbf{#1}}

\title{CAFE: Confining Feature Distillation to Language-Routed Adapters for Compact Multilingual Automatic Speech Recognition}

\author{
 \textbf{Tianchi Liu},
 \textbf{Nicholas Moratelli},
 \textbf{Khurram Azeem Hashmi},
\\
 \textbf{Mohammadreza Zolfaghari}, 
 \textbf{Rishabh Jain},
 \textbf{Louis Lu},
 \textbf{Changdae Park},
 \textbf{Amril Nazir}
\\
\\
 Tether AI Research
}

\begin{document}

\maketitle

\begin{abstract}
    Feature distillation is usually expected to add to logit distillation. We find that this expectation does not hold for two common feature losses in compact multilingual automatic speech recognition (ASR), and analyze why through their gradients. The gradient of a projection-based loss is orthogonal to the task gradient, while that of Centered Kernel Alignment (CKA), a similarity-based loss, is weakly aligned with it. CKA lowers the word error rate (WER) on non-English languages, but at the expense of data-rich English. Three loss-level remedies (one trained, two probed) all fall short, and the CKA gradient overlaps with the subspace English uses in the shared encoder by 97.5\% to 99.5\% (about 81\% by chance). Guided by this diagnosis, we propose CAFE (\textbf{C}onfined-\textbf{A}dapter \textbf{Fe}ature distillation), which confines the CKA gradient to adapters that English bypasses, addressing the trade-off in the parameters rather than in the loss. On the Open ASR Leaderboard, CAFE lowers the average WER from 8.55\% to 8.02\%, beats an adapter-only capacity control at 8.51\%, and makes English more stable across checkpoints. At 254M parameters, it trails the 0.6B teacher by 2 WER points and is competitive with much larger models across 48 test sets in 24 languages. We use 564k hours of training data and state-of-the-art models to keep the findings relevant to practice.\footnote{Checkpoints and usage instructions are available at \url{https://huggingface.co/qvac/TranscribePsy-250M-EU}. TranscribePsy-250M-EU is the 25-language CAFE model.}
\end{abstract}

\section{Introduction}
\label{sec:intro}

\begin{figure}[t]
    \centering
    \includegraphics[width=\columnwidth]{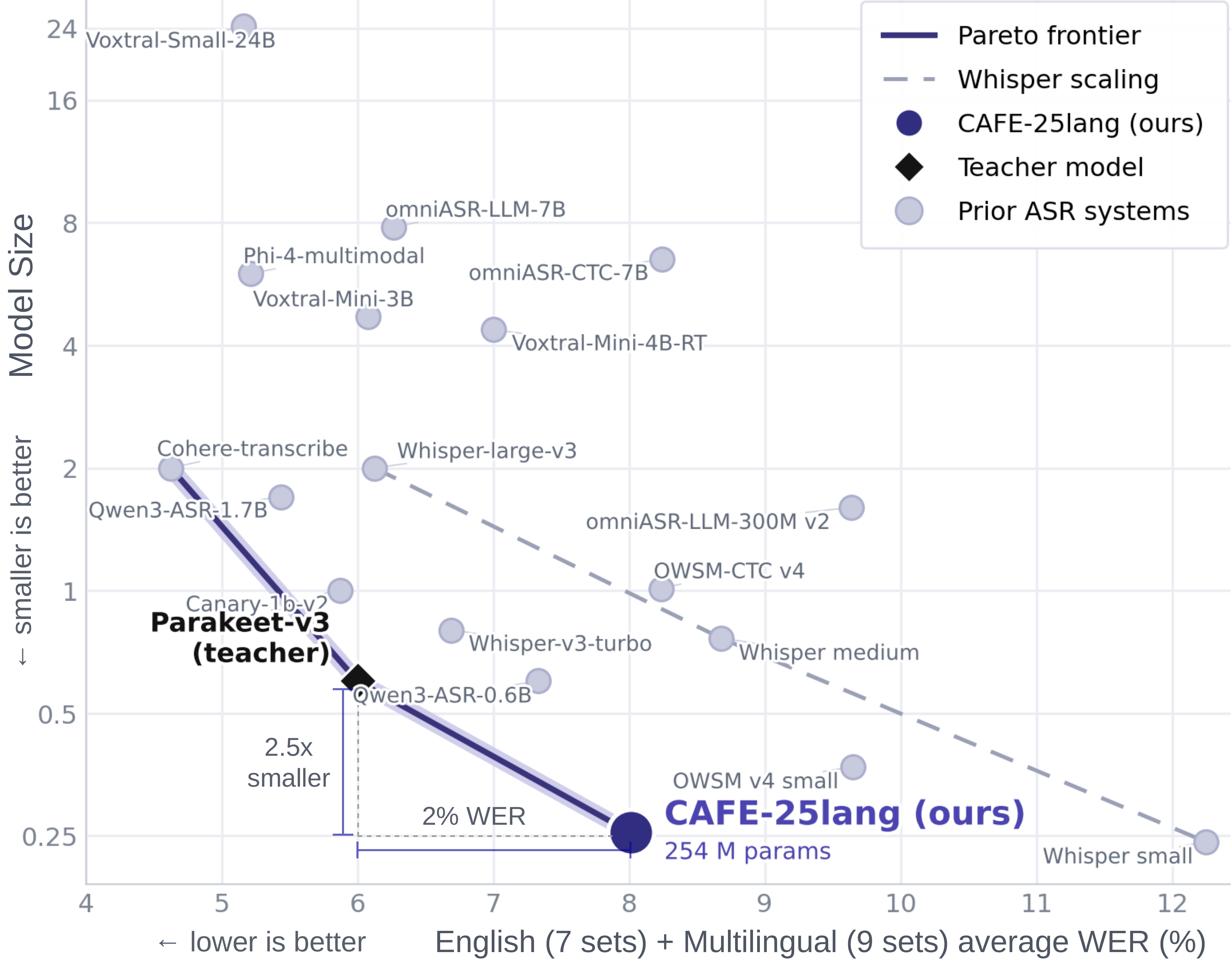}
    \caption{Averaged English and Multilingual WER vs. model size on the Open ASR Leaderboard.}
    \label{fig:pareto}
    \vspace{-5 mm}
\end{figure}

Automatic speech recognition (ASR) increasingly calls for one model that transcribes many languages, letting high-resource languages transfer to low-resource ones \citep{radford2023whisper,pratap2024mms}.
Such a model must serve data-rich English and a long tail of lower-resource languages at the same time, and the gap between them persists \citep{wang2025speechbt,chen2025metamixspeech,xiao-etal-2026-adapting}. Many applications also need the model to run on device, for privacy, low latency, and offline use \citep{jeffries2024moonshine,kudlur2026moonshinev2,li2025power,kamahori2025liteasr,hashmi2026visionpsy}. A compact student performs better when initialized by structured pruning of a strong teacher rather than at random \citep{zhang2025prunernnt,lin2026arkasr} and trained with logit distillation from it \citep{hinton2015kd,gandhi2023distilwhisper}.

Many works also match intermediate features \citep{romero2015fitnets,zagoruyko2017attention,park2019rkd,tian2020crd}, and speech encoders are often compressed this way \citep{lee2024colld,10626978,chi25_interspeech}, on the assumption that a feature loss adds useful signal beyond the logits. In a multilingual model, however, all languages share one encoder, and a signal that helps some languages can hurt others \citep{chang2024curse}. Whether feature distillation still adds to logit distillation in this setting, and why, is not well understood.

This paper first diagnoses why feature distillation fails in this setting, and then uses the diagnosis to fix it.  We distill a compact multilingual ASR student, initialized by structured pruning of the teacher, and find that neither of two common feature losses adds to logit distillation: a projection-based loss has almost no effect, and Centered Kernel Alignment (CKA) \citep{cortes2012cka,kornblith2019cka}, a similarity-based loss, trades English for the non-English languages. 
To find out why, we analyze the gradient that each loss sends to the encoder. The projection-based gradient is orthogonal to the task, while the CKA gradient acts on the same encoder directions that English uses. Guided by this diagnosis, we propose \textbf{C}onfined-\textbf{A}dapter \textbf{Fe}ature distillation (CAFE), which applies the CKA loss only through adapters that English bypasses and stops its gradient from reaching the shared encoder. 
Our contributions are:

\begin{itemize}
\item \textbf{Analysis.} We show that two common feature losses do not add to logit distillation in compact multilingual ASR, and use their gradients to explain why (Sections~\ref{sec:live} and~\ref{sec:conditional}).
\item \textbf{Diagnosis.} The multilingual CKA gradient overlaps with the subspace English uses by 97.5--99.5\% (about 81\% by chance), which explains why masking English in the loss fails and argues against two other loss-level remedies (Section~\ref{sec:whyfail}).
\item \textbf{Method.} CAFE separates the trade-off in the parameters rather than in the loss, so feature distillation no longer comes at the expense of English (Section~\ref{sec:method}).
\item \textbf{Results.} CAFE's gains over the baselines come mainly from English, with a smaller multilingual gain. It beats a capacity control with the same adapters but no feature loss, and halves the variation of English WER across checkpoints. At 254M parameters it is 2 WER points behind the 0.6B teacher (Section~\ref{sec:exp}).
\end{itemize}

\section{Related Work}
\label{sec:related}

\paragraph{Logit and feature distillation.}
Knowledge distillation matches a student's outputs to a teacher's \citep{hinton2015kd}. Feature-based variants add intermediate signal by matching activations \citep{romero2015fitnets,zagoruyko2017attention} or relations among examples \citep{park2019rkd,tian2020crd}, and a parallel line improves the logit term \citep{zhao2022dkd,huang2022dist}. It is commonly assumed that a feature term helps on top of a logit term. In our setting, we find this does not hold for two common feature losses.

\paragraph{Representational similarity and CKA.}
CKA \citep{cortes2012cka} became a standard tool for comparing neural representations across models of different width and initialization \citep{kornblith2019cka}. Recent work also uses it as a training loss between models of different dimensions, with positive results, including in multilingual settings \citep{dasgupta2025cka}. Other feature distillation methods, such as VkD \citep{miles2024vkd}, project the student into the teacher space, or match layers with a contrastive loss to compress multilingual speech encoders \citep{lee2024colld}.
We use similar losses but reach different conclusions, mainly because of the setting: a compact multilingual ASR student initialized by pruning, trained with strong logit distillation, on large-scale data dominated by English. Here, a VkD-style projection loss has almost no effect, and CKA helps only the languages with fewer resources compared with English. We limit these claims to multilingual ASR.

\paragraph{Multilingual ASR distillation.}
Distillation is a standard way to compress speech models, through pseudo labels \citep{gandhi2023distilwhisper,waheed2025udistil} or dual-mode multilingual encoders \citep{male2025durep}. \citet{dellalibera2024clmasr} benchmark adding languages to a multilingual model without catastrophic forgetting. Closest to us, \citet{ferraz2024distilwhisper} add language-specific gated experts to a distilled Whisper~\cite{radford2023whisper} student to raise per-language accuracy. Our adapters serve a different purpose. They are not there to add capacity, but to keep a feature distillation gradient inside the adapters and away from the shared encoder.

\paragraph{Adapters and conditional computation.}
Bottleneck adapters \citep{houlsby2019adapters} and low-rank updates~\cite{hu2022lora} adapt a frozen backbone, and language-specific adapters improve multilingual performance without disturbing other languages \citep{bapna2019adapters,pfeiffer2020madx}. Recent work selects language-specific LoRA experts with the model's own language identification \citep{kwok2025twostagelora, li25p_interspeech}, or studies bottleneck adapters for low-resource ASR \citep{samin2025convadapt, moe_accent, lee-etal-2026-mixture, li-etal-2026-lcma, oukid-etal-2026-puma}. In these works the adapter is used for transfer, while we use it to restrict where a distillation gradient can act.

\paragraph{When distillation helps or hurts.}
Matching the teacher more closely does not always improve generalization \citep{stanton2021doeskd}, and a stronger teacher can produce a worse student \citep{busbridge2025scalinglaws}. Our finding is in the same spirit, applied to feature matching. When a feature loss conflicts with the task, gradient surgery projects the conflicting gradients apart \citep{yu2020pcgrad, huang2025deepkd}. In our case, this would 
barely help, as the gradients are orthogonal rather than opposed. This is why we separate them in the parameters.

\section{Preliminaries}
\label{sec:objectives}

We distill Parakeet-TDT-0.6B-v3~\cite{parakeet, tdt2023, rekesh2023fastconformer} into a 254M student, initialized by structured pruning of the teacher~\cite{NEURIPS2024_48229913}. We report WER on the Open ASR Leaderboard \citep{srivastav2025openasr}: EN-7 and ML-9 summarize its seven English and nine multilingual sets. We evaluate each system with checkpoint averaging (CA)~\cite{polyak1992acceleration, vaswani2017attention}, which averages the weights of its last 7 checkpoints. Section~\ref{sec:setup} gives the full details.

The training objective combines a supervised transducer loss $\mathcal{L}_{\text{sup}}$, logit distillation $\mathcal{L}_{\text{logit}}$, and an optional feature distillation loss $\mathcal{L}_{\text{feat}}$,
\begin{equation}
\mathcal{L} = \mathcal{L}_{\text{sup}} + \lambda_{\text{logit}}\,\mathcal{L}_{\text{logit}} + \lambda_{\text{feat}}\,\mathcal{L}_{\text{feat}},
\end{equation}
where the $\lambda$ terms are loss weights. The logit term is the KL divergence between student and teacher label and duration distributions at temperature $\tau$,
\begin{equation}
\mathcal{L}_{\text{logit}} = \tau^2 \sum_{c} p^{T}_{c}\big(\log p^{T}_{c} - \log p^{S}_{c}\big),
\end{equation}
summed over the label and duration heads. We call the recipe with $\lambda_{\text{feat}}{=}0$ the Baseline.

\begin{figure*}[h]
    \centering
    \includegraphics[width=0.99\textwidth]{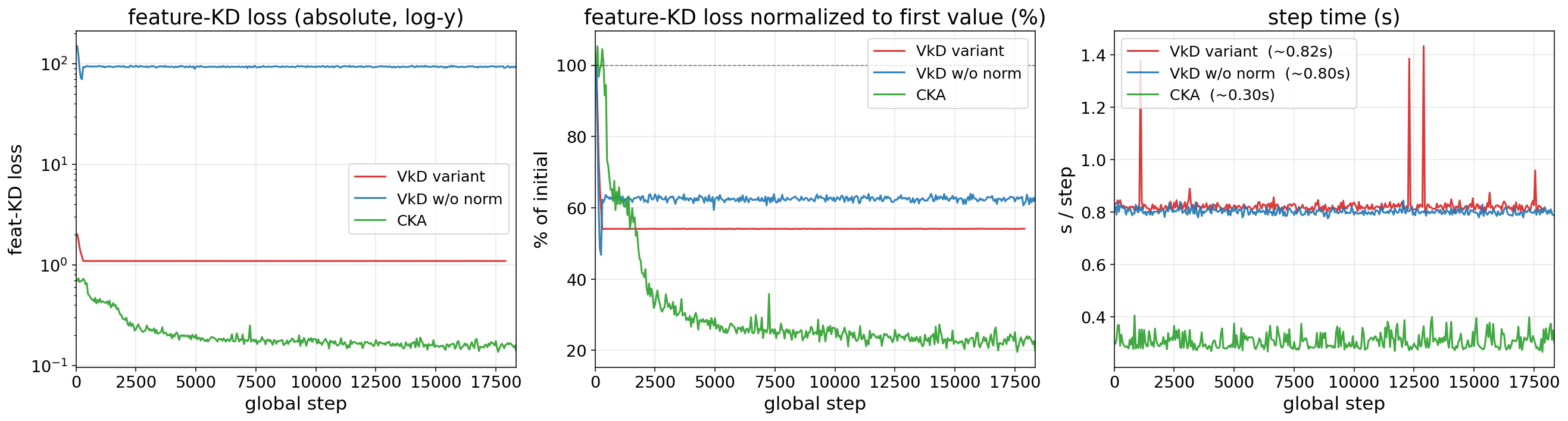}
    \vspace{-3 mm}
    \caption{Feature-only runs: the same supervised recipe plus one feature distillation loss, with logit distillation off. (a) Feature loss (log scale). (b) The same loss divided by its initial value. (c) Wall-clock time per training step. VkD w/o norm is the projection loss without LayerNorm.}
    \label{fig:probe}
    \vspace{-3 mm}
\end{figure*}

Given student features $S\in\mathbb{R}^{N\times d_s}$ and teacher features $T\in\mathbb{R}^{N\times d_t}$ over $N$ valid frames ($d_s{=}640$, $d_t{=}1024$), we compare two types of loss for $\mathcal{L}_{\text{feat}}$. The first is a VkD-style projection loss \citep{miles2024vkd}, called VkD variant in our tables. A projection $P$ with orthonormal columns ($P^\top P{=}I$) maps the student to the teacher dimension, $\hat S{=}S P^\top$, and a LayerNorm without affine parameters ($\mathrm{LN}$) is applied to both $\hat S$ and the teacher before matching them frame by frame,
\begin{equation}
\mathcal{L}_{\text{VkD}} = \frac{1}{N}\sum_{n}\tfrac{1}{d_t}\big\|\tilde S_n - \tilde T_n\big\|_2^2,
\end{equation}
with $\tilde S=\mathrm{LN}(\hat S)$ and $\tilde T=\mathrm{LN}(T)$; the w/o norm variant omits the LayerNorm. The second is linear CKA \citep{cortes2012cka,kornblith2019cka}, which compares the similarity structure of the two feature sets, needs no projection, and works with $d_s\neq d_t$,
\begin{equation}
\mathcal{L}_{\text{CKA}} = 1 - \frac{\|\bar S^\top \bar T\|_F^2}{\|\bar S^\top \bar S\|_F\,\|\bar T^\top \bar T\|_F},
\label{eq:cka}
\end{equation}
where $\bar S, \bar T$ are column centered, and the loss is averaged over the distilled layers (Section~\ref{sec:setup}).

\section{Analysis: When and Why Feature Distillation Helps or Hurts}
\label{sec:analysis-motiv}

\subsection{Which feature loss actually trains the student?}
\label{sec:live}

We first check whether each feature loss changes the student at all. We run three feature-only probes with logit distillation off, identical except for the feature loss, so the loss curve shows directly whether the optimizer can reduce it.

\paragraph{The projection loss stalls, while CKA keeps decreasing.} Figure~\ref{fig:probe} shows the feature loss and the step time. The projection loss, with or without LayerNorm, stops decreasing right after warmup, while CKA keeps falling, to about a quarter of its initial value (panel b). This is not specific to the probes: in the full recipe with logit distillation, the projection loss stays flat from step 49 to the end of training.

\begin{table}[h]
\centering\small
\resizebox{0.8\columnwidth}{!}{%
\begin{tabular}{lcc}
\toprule
feature term & $\|g_{\text{feat}}\|/\|g_{\text{sup}}\|$ & $\cos(g_{\text{feat}},g_{\text{sup}})$ \\
\midrule
VkD variant & $7.3\times10^{-4}$ & $+0.009$ \\
VkD w/o norm & $0.15$ & $-0.005$ \\
CKA & $1.1\times10^{-3}$ & \best{$+0.087$} \\
\bottomrule
\end{tabular}
}
\vspace{-1 mm}
\caption{Median ratio of gradient norms, and cosine similarity between the feature and supervised gradients.}
\label{tab:grad}
\vspace{-3 mm}
\end{table}

\paragraph{Gradient size and direction.} On real batches, we measure the gradient each feature loss sends to the encoder and compare it with the supervised gradient (Table~\ref{tab:grad}). The gradients of the VkD variant and of CKA are both about $10^{3}$ times smaller than the supervised gradient, but they differ in direction. Both VkD gradients, with or without LayerNorm, are orthogonal to the task (cosine near zero), while CKA is the only feature loss with a consistent, weakly positive alignment, with a cosine of about $0.09$. This small aligned part is what makes CKA trainable. The much larger orthogonal part is the structure that CKA adds to the representation, and Section~\ref{sec:conditional} shows that it helps some languages and hurts others.

\paragraph{Ruling out limited capacity.} One explanation for the stalled projection loss is that a 640-dimensional student cannot represent the 1024-dimensional teacher features. This is not the case. The teacher features are low rank, and 640 dimensions capture about 99\% of their energy across five languages. The optimizer simply does not follow this gradient. Consistent with this, the CKA between student and teacher features rises from 0.49 to between 0.83 and 0.87 under all three losses. The two VkD versions reach similar values although their gradients differ in size by about 200 times, so for VkD this alignment comes from the task loss rather than the feature loss.

\paragraph{Cost.} The projection is also expensive. The standard orthogonal parametrization rebuilds an orthonormal matrix in every forward and backward pass, which raises the feature cost per step by about $230\times$, from 1.76 to 407 ms for six layers. CKA needs no projection, and a training step with CKA is about $2.8\times$ faster than with the projection loss (Figure~\ref{fig:probe}c). It is therefore both the more effective and the cheaper choice, and we use it from here on.

\subsection{CKA helps non-English languages and hurts English}
\label{sec:conditional}

We train a full run, called CKA in Table~\ref{tab:lossvariants}, that is identical to the Baseline except for a CKA term with $\lambda_{\text{feat}}{=}32$. We chose this weight with a small sweep on a held-out set, not on the test sets. A large fixed weight works because the CKA loss shrinks as alignment improves; the VkD variant, in contrast, needed a decaying weight.

\begin{table}[t]
    \centering\footnotesize
    \begin{tabular}{l c c c}
    \toprule
    System & EN-7 & ML-9 & Avg \\
    \midrule
    Baseline & 9.85 & 7.24 & \textbf{8.55} \\
    VkD variant & 10.18 & 7.22 & 8.70 \\
    CKA & 11.25 & \textbf{6.83} & 9.04 \\
    Lang-gated CKA & 11.21 & 7.18 & 9.19 \\
    \bottomrule
    \end{tabular}
    \vspace{-1mm}
    \caption{Feature losses on top of the Baseline (WER after CA). Lang-gated CKA applies CKA to non-English frames only.}
    \label{tab:lossvariants}
    \vspace{-3mm}
\end{table}

CKA gives the best multilingual WER of all systems, 6.83\% on ML-9, but the worst English WER, 11.25\% on EN-7 (Table~\ref{tab:lossvariants}). The English degradation is significant (Appendix~\ref{app:sig}) and appears late in training: English follows the Baseline early on and then drifts upward. English WER is noisy across checkpoints for every system (Section~\ref{sec:exp}), but CKA adds a consistent upward shift.

We attribute this to the part of the CKA gradient that is orthogonal to the task (Section~\ref{sec:live}). On the non-English languages, where the student underfits, the teacher's representation structure is a useful prior. On English, abundant data and logit distillation already give a strong signal, so this structure adds little, and its task-orthogonal part pulls the shared encoder away from what English needs. English is not at its limit at this size: the student still trails the teacher, and CAFE improves it once this pull is removed (Section~\ref{sec:exp}). Whether feature distillation helps therefore depends on the language.

\begin{figure*}[t]
    \centering
    \includegraphics[width=0.95\textwidth]{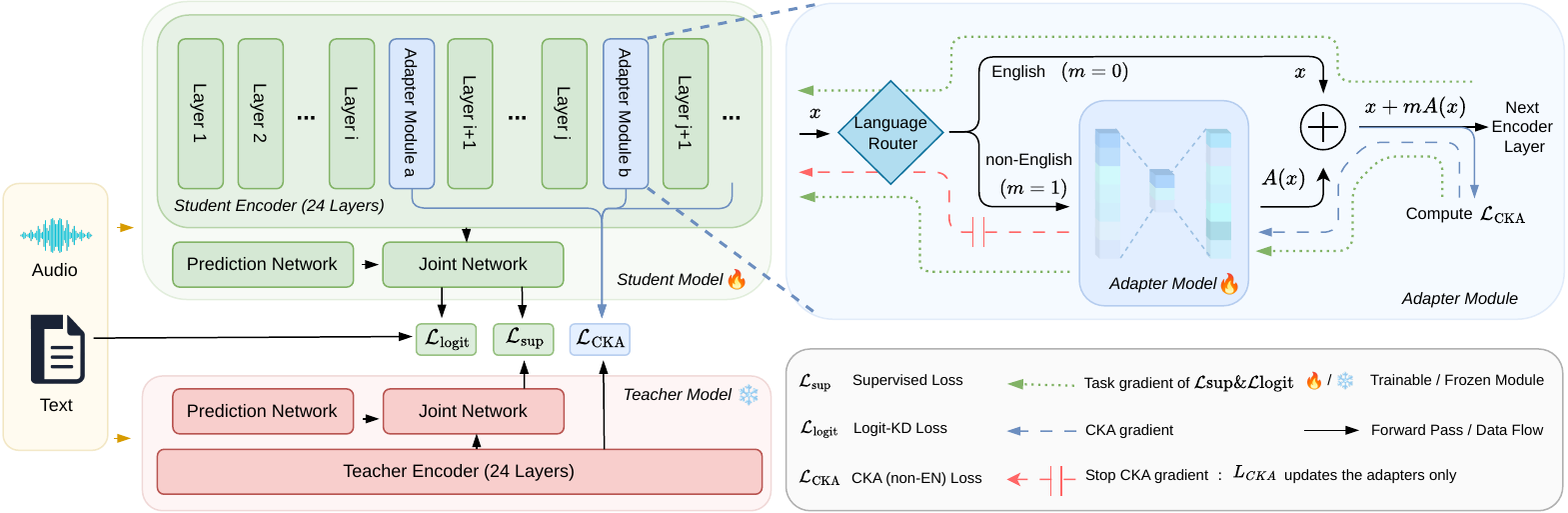}
    \caption{CAFE: feature distillation confined to adapters. There is one adapter at each distilled layer $\ell\in\{3,7,11,15,19,23\}$, and only two are drawn. $A(\cdot)$ denotes the residual update added by the adapter.}
    \label{fig:method}
    \vspace{-3 mm}
\end{figure*}

\subsection{Why the trade-off is hard to fix at the loss level}
\label{sec:whyfail}

If the gain and the harm were separable, a targeted change to the loss could keep the multilingual gain and drop the English harm. We examine three such changes, one in a training run and two with training-free probes. The trained one does not help, and the probes argue against gating and projection, pointing to the shared encoder as the cause.

Multilingual data dominates the CKA term. By sampling weight, English makes up 13.6\% of training frames, 5.2\% human-labeled and 8.4\% pseudo-labeled, and the other languages 86.4\%, mostly pseudo-labeled VoxPopuli. The CKA statistics are therefore computed mostly on non-English frames.

\paragraph{Remedy 1: Masking English in the loss.} Lang-gated CKA uses the true language labels to apply CKA to non-English frames only. However, as shown in Table~\ref{tab:lossvariants}, English WER stays as high as with plain CKA, at 11.21\% versus 11.25\%, and multilingual WER rises from 6.83\% to 7.18\%. Removing English frames from the loss does not stop the multilingual CKA gradient from changing the shared encoder weights that English also uses. It also discards English frames whose alignment signal was useful.

\paragraph{Remedy 2: Gating by the loss gap.} A language-agnostic option is to switch CKA off wherever the student has caught up with the teacher, measured by the loss gap $\Delta=\text{nll}_S-\text{nll}_T$. A training-free probe on about 4000 utterances shows that the gap is positive almost everywhere (96\% to 98\% of utterances) and is not larger on English than on other languages. This holds on human-labeled data and late in training, when English is already degrading. The student is still behind the teacher on almost every utterance, so the task loss does not reveal the harm. The harm lies in the structure of the features.

\paragraph{Remedy 3: Subspace projection.} A third option is to remove from the CKA gradient its component in the subspace that English uses. For each distilled layer, we measure the share of the multilingual CKA gradient energy that falls inside the top-$k$ subspace of English activations,
\begin{equation}
\text{overlap}(k) = \frac{\operatorname{tr}(U_k^\top G\, U_k)}{\operatorname{tr}(G)},
\end{equation}
where $U_k$ spans the top-$k$ principal directions of English activations and $G$ is the second moment of the per-frame CKA gradient. English activations spread over most of the 640 dimensions, and the CKA gradient overlaps with their subspace by 97.5\% to 99.5\%, far above the chance level of about 81\%. Projecting this part out would keep only a few percent of the CKA gradient, which amounts to removing CKA. The overlap stays stable during training. The multilingual gain and the English harm use the same encoder directions.

All three run into the same problem: all languages share one representation, and the CKA gradient comes mostly from multilingual data. The simplest fix consistent with this is to give the non-English side its own parameters. Lowering the CKA weight late in training is not a solution either, because nothing in the loss tells us when to do it.

\section{Proposed CAFE: Confined-Adapter Feature Distillation}
\label{sec:method}

Since the trade-off is hard to separate in the loss, CAFE separates it in the parameters. It routes non-English utterances through a shared set of adapters and applies the feature loss only through them, so the feature gradient never reaches the shared encoder that English relies on, as shown in Figure~\ref{fig:method}. The CKA gradient thus reaches only the adapters, which only non-English utterances use.

\paragraph{Language-routed adapters.} At each distilled layer $\ell$ we insert one residual bottleneck adapter, shared by all non-English languages, with update
\begin{equation}
A_\ell(x) = U_\ell\,\phi(D_\ell\,x),
\end{equation}
with down projection $D_\ell\in\mathbb{R}^{r\times d}$, up projection $U_\ell\in\mathbb{R}^{d\times r}$, GELU nonlinearity $\phi$, and rank $r$. $U_\ell$ is initialized to zero, so $A_\ell$ starts at zero and the model at initialization is identical to the Baseline. The adapter is applied only to non-English utterances, through a per-utterance mask $m\in\{0,1\}$,
\begin{equation}
\tilde x = x + m\,A_\ell(x),
\end{equation}
so the outputs for English are bit-identical to those of the encoder without adapters. The adapters are the only language-dependent parameters, and the encoder remains a single multilingual model.

CAFE keeps the base recipe of Section~\ref{sec:objectives} and changes only the feature term. CKA is computed between the adapted non-English features and the teacher features, with the same fixed $\lambda_{\text{feat}}$. No schedule is needed, because the CKA loss shrinks as alignment improves.

\paragraph{Gradient confinement.} The CKA term trains the adapters to match the teacher, and a stop-gradient keeps its gradient out of the shared encoder, as shown in the blue panel of Figure~\ref{fig:method},
\begin{equation}
\Phi^{\text{CKA}}_\ell = \operatorname{sg}(x) + A_\ell(\operatorname{sg}(x)),
\label{eq:sg}
\end{equation}
where $\operatorname{sg}$ is the stop-gradient operator. The supervised and logit losses update the shared encoder as usual. Confining the CKA gradient in this way has two consequences. First, the CKA term cannot affect English directly at any $\lambda_{\text{feat}}$, because English bypasses the adapters and the shared encoder receives no CKA gradient. Second, the CKA weight does not need to be lowered late in training, because the task gradient keeps the adapters close to the task optimum. To account for the added parameters, we also train a control named Capacity that has the same adapters but no feature loss.

\section{Experimental Setup}
\label{sec:setup}

\paragraph{Training.} The student encoder has 24 layers of dimension 640, against 1024 for the teacher\footnote{\url{https://huggingface.co/nvidia/parakeet-tdt-0.6b-v3}}. 
All runs share the same data, optimizer, and schedule: AdamW with Noam annealing (base scale 3.0, peak learning rate ${\approx}1.2\times10^{-3}$, 10k warmup steps) for 350k steps on NVIDIA H100. Batches are formed dynamically by duration with up to 22\,s of audio per GPU, with 3 gradient accumulation steps.
Logit distillation uses $\lambda_{\text{logit}}{=}5$ and $\tau{=}2$. The feature distillation is applied at layers ${\{3,7,11,15,19,23\}}$. The CAFE adapters have rank $r{=}64$.

\paragraph{Data.} The student is trained on about 116.9 million utterances (564k hours) in 25 languages from 10 corpora, mostly VoxPopuli, YODAS, Multilingual LibriSpeech (MLS), and Libri-Light~\citep{wang2021voxpopuli,li2023yodas,pratap2020mls,kahn2020librilight}. VoxPopuli, YODAS, and Libri-Light use the Granary pseudo-labels~\citep{raokoluguri25_interspeech}. English is the largest language by hours (35\%), but sampling reduces it to 14\% of training frames. We evaluate on 48 held-out test sets (296k utterances, 24 languages): the public Open ASR Leaderboard\footnote{\url{https://huggingface.co/spaces/hf-audio/open_asr_leaderboard}}, whose EN-7 sets are English short-form and whose ML-9 sets come from FLEURS (5) and MLS (4), plus the larger multilingual Common Voice v25~\citep{ardila2020commonvoice} and Multilingual LibriSpeech (MLS-8) test sets. Detailed data statistics are in Appendix~\ref{app:data}.

\paragraph{Model selection.} Checkpoints are saved every 6k steps, and CA averages the last 7. The ranking holds for windows of 5, 7, and 10 checkpoints (Table~\ref{tab:nrobust}). CA smooths the fluctuation between consecutive checkpoints at no extra training cost. It is widely used in ASR~\cite{shinayama26_interspeech, zhou-etal-2024-copyne, gaido-etal-2024-mosel, lam-etal-2025-prepending, chen-etal-2024-m3av} and other speech-related tasks~\cite{alastruey-etal-2024-unveiling, 11463369, 11222612, zhu-etal-2025-zipa}. 

\paragraph{Systems.} We compare four 25-language systems in a $2\times2$ design, with or without the adapters and the CKA term (Table~\ref{tab:full}). Baseline and CKA are the systems of Section~\ref{sec:analysis-motiv}, and Capacity has the same adapters as CAFE but no feature loss. CAFE versus Capacity thus isolates the confined CKA term, and Capacity versus Baseline isolates the adapters alone. CAFE w/o sg is identical to CAFE but without the stop-gradient, to test its effect. Baseline-8lang and CAFE-8lang are trained on the 8 languages of MLS, with a different seed.

\section{Results}
\label{sec:exp}

\begin{table*}[t]
    \centering
    \resizebox{0.90\textwidth}{!}{%
    \begin{tabular}{l c c l HH HH GG}
    \toprule
     & & & & \multicolumn{2}{c}{English (7 sets)} & \multicolumn{2}{c}{Multilingual (9 sets)} & \multicolumn{2}{c}{Avg (EN, ML)} \\
    \cmidrule(lr){5-6}\cmidrule(lr){7-8}\cmidrule(lr){9-10}
    system & Lang. & Adapter & feature-KD & mean\,$\pm$\,std & CA & mean\,$\pm$\,std & CA & mean\,$\pm$\,std & CA \\
    \midrule
    Baseline & 25 & -- & -- & 10.33\,$\pm$\,0.82 & 9.85 & 7.59\,$\pm$\,0.20 & 7.24 & 8.96\,$\pm$\,0.39 & 8.55 \\
    CKA & 25 & -- & CKA & 11.42\,$\pm$\,0.67 & 11.25 & \best{7.22\,$\pm$\,0.12} & \best{6.83} & 9.32\,$\pm$\,0.33 & 9.04 \\
    Capacity & 25 & \checkmark & -- & 9.98\,$\pm$\,0.53 & 9.67 & 7.77\,$\pm$\,0.11 & 7.34 & 8.88\,$\pm$\,0.27 & 8.51 \\
    CAFE w/o sg & 25 & \checkmark & CKA (no sg) & 10.75\,$\pm$\,0.69 & 10.44 & 7.37\,$\pm$\,0.12 & 6.95 & 9.06\,$\pm$\,0.32 & 8.70 \\
    \textbf{CAFE (ours)} & 25 & \checkmark & CKA (confined) & \best{9.32\,$\pm$\,0.32} & \best{9.03} & 7.43\,$\pm$\,0.13 & 7.00 & \best{8.38\,$\pm$\,0.19} & \best{8.02} \\
    \midrule
    \midrule
    Baseline-8lang & 8 & -- & -- & 10.32\,$\pm$\,0.54 & 10.11 & 7.08\,$\pm$\,0.15 & 6.62 & 8.70\,$\pm$\,0.22 & 8.36 \\
    \textbf{CAFE-8lang (ours)} & 8 & \checkmark & CKA (confined) & \best{8.93\,$\pm$\,0.27} & \best{8.71} & \best{6.91\,$\pm$\,0.12} & \best{6.53} & \best{7.92\,$\pm$\,0.13} & \best{7.62} \\
    \bottomrule
    \end{tabular}%
    }
    \vspace{-2 mm}
    \caption{Main comparison: WER (lower is better) on the Open ASR Leaderboard sets \citep{srivastav2025openasr}. \emph{Lang.}\ is the number of training languages. Each system differs from the Baseline (supervised loss plus logit distillation) in two factors: \emph{Adapter} (whether it adds the language-routed adapters) and \emph{feature-KD} (the feature loss: \emph{--} for none, and \emph{CKA (confined)} for CKA confined to the adapters). \emph{CAFE w/o sg} is CAFE without the stop-gradient of Eq.~\eqref{eq:sg}. For each system we give the mean$\pm$std over the last 7 checkpoints and the WER of their checkpoint average (CA).}
    \label{tab:full}
    \vspace{-1 mm}
\end{table*}

\begin{table}[t]
    \centering\footnotesize
    \resizebox{0.82\columnwidth}{!}{%
    \begin{tabular}{l r r}
    \toprule
     & \textbf{CAFE} & Baseline \\
    \midrule
    Params & 254.45M & 253.95M \\
    RTFx bs1 $\uparrow$ & 212.9 $\pm$ 6.9 & 221.3 $\pm$ 5.2 \\
    RTFx bs32 $\uparrow$ & 2602.3 $\pm$ 68.9 & 2656.6 $\pm$ 79.8 \\
    P50 lat. (s) $\downarrow$ & 0.089 $\pm$ 0.001 & 0.088 $\pm$ 0.000 \\
    Peak mem (MB) $\downarrow$ & 2593.4 & 2592.5 \\
    \bottomrule
    \end{tabular}
    }
    \vspace{-1mm}
    \caption{Adapter overhead: CAFE versus the same model without adapters (Baseline). RTFx is the inverse real-time factor, P50 is the median latency at batch size 1, and peak memory is measured at batch size 32.}
    \label{tab:efficiency}
    \vspace{-3mm}
\end{table}

\paragraph{Main result.}
CAFE improves on the Baseline in English and multilingual WER: EN-7 falls from 9.85\% to 9.03\%, and ML-9 from 7.24\% to 7.00\%. The checkpoint means show the same order, so the gain is not a product of averaging. CAFE also beats the Capacity control on both, which reaches 9.67\% on EN-7 and 7.34\% on ML-9, and both English gains are significant (Appendix~\ref{app:sig}). The adapters alone give only a small English gain, so the improvement comes from the confined CKA term rather than from the extra parameters. Plain CKA has the lowest multilingual WER but the worst English. Against Capacity, the confined CKA term still lowers ML-9, nearly as much as plain CKA gains over the Baseline. 

\paragraph{Effect of the stop-gradient.} Removing the stop-gradient brings the English degradation back. CAFE w/o sg reaches 10.44\% on EN-7, worse than the Baseline, while its ML-9 of 6.95\% matches that of CAFE. Its English standard deviation across checkpoints is 0.69, close to that of plain CKA and twice that of CAFE. The stop-gradient thus protects English at almost no multilingual cost.

\paragraph{Replication on 8 languages.} We train a matched pair, Baseline-8lang and CAFE-8lang, on the 8 languages covered by MLS, otherwise identical to their 25-language counterparts. The results in Table~\ref{tab:full} show the same pattern with less data: EN-7 WER is improved from 10.11\% to 8.71\%, and ML-9 from 6.62\% to 6.53\%. The effect is therefore not specific to the 25-language setup.

\paragraph{Adapter overhead.} Table~\ref{tab:efficiency} compares CAFE with the same model without adapters. The adapters lower throughput by about 2.0\%--3.8\% and leave latency and memory almost unchanged, so CAFE runs as fast as a standard transducer of its size. All efficiency numbers follow the single-GPU protocol in Appendix~\ref{app:efficiency}.

\paragraph{Stability.} English WER varies across checkpoints for every 25-language system, but least for CAFE, with a standard deviation of 0.32, and most for the Baseline, with 0.82 as shown in Table~\ref{tab:full}. Confining CKA to the adapters both lowers English WER and more than halves this variation. As a result, CAFE's WER changes little with the CA window (ablation in Appendix~\ref{app:window}).

\begin{table*}[t]
\centering\small
\setlength{\tabcolsep}{4pt}
\renewcommand{\arraystretch}{1.05}
\resizebox{1\textwidth}{!}{%
\begin{tabular}{@{}l r !{\vrule width 0.6pt} r r r r r !{\vrule width 0.6pt} r r r r !{\vrule width 0.6pt} r@{}}
\toprule
 & \multicolumn{1}{c}{\makecell{Parakeet\\(teacher)}} & \multicolumn{1}{c}{\makecell{Whisper\\medium}} & \multicolumn{1}{c}{\makecell{omniASR-LLM\\300M v2}} & \multicolumn{1}{c}{\makecell{OWSM-\\CTC v4}} & \multicolumn{1}{c}{\makecell{Qwen3-ASR\\0.6B}} & \multicolumn{1}{c}{\makecell{omniASR-\\CTC-1B v2}} & \multicolumn{1}{c}{\makecell{Whisper\\small}} & \multicolumn{1}{c}{\makecell{omniASR-CTC\\300M v2}} & \multicolumn{1}{c}{\makecell{OWSM\\v4 small}} & \multicolumn{1}{c}{\makecell{\textbf{CAFE}\\\textbf{(Ours)}}} & \multicolumn{1}{c}{\makecell{\textbf{CAFE-}\\\textbf{8lang}}} \\
\midrule
Params (M) & 627 & 764 & 1{,}632 & 1{,}011 & 782 & 976 & 242 & 326 & 370 & 254 & 254 \\
Number of languages & 25 & $\sim$100 & $>$1600 & 75 & 30 & $>$1600 & $\sim$100 & $>$1600 & 75 & 25 & 8 \\
RTFx bs1 $\uparrow$ & 198 & 21 & 10 & 590 & 18 & 513 & 35 & 848 & 39 & 213 & 213 \\
RTFx bs32 $\uparrow$ & 2{,}520 & 386 & 117 & 1{,}282 & 310 & 2{,}142 & 546 & 3{,}405 & 37 & 2{,}602 & 2{,}602 \\
P50 lat.\ bs1 (s) $\downarrow$ & 0.095 & 0.704 & 1.990 & 0.033 & 1.070 & 0.037 & 0.417 & 0.024 & 0.501 & 0.089 & 0.089 \\
Peak mem (MB) $\downarrow$ & 3{,}310 & 6{,}932 & 17{,}310 & 25{,}803 & 6{,}266 & 16{,}033 & 2{,}617 & 14{,}802 & 2{,}505 & 2{,}593 & 2{,}593 \\
\midrule
\textit{Open ASR Leaderboard}\\
\ \ EN short-form (7) & 6.83 & 11.21 & 11.28 & \underline{7.90} & \textbf{6.50} & 12.21 & 13.91 & 17.94 & \textbf{8.73} & \underline{9.03} & 8.71 \\
\ \ Multilingual (9) & 5.14 & \textbf{6.36} & 7.58 & 8.74 & 8.29 & \underline{7.16} & 10.76 & 12.68 & \underline{10.71} & \textbf{7.00} & 6.53 \\
\ \ EN long-form (4) & 10.77 & 16.28 & 23.50 & \underline{12.22} & \textbf{10.91} & 54.82 & 16.71 & 42.45 & \textbf{13.01} & \underline{13.35} & 12.99 \\
\midrule
MLS-8 (8) & 6.66 & \underline{7.76} & 7.79 & 10.65 & 12.60 & \textbf{7.71} & \underline{11.51} & 11.89 & 11.84 & \textbf{6.58} & 6.25 \\
Common Voice v25 (24) & 16.48 & 26.94 & \textbf{17.72} & 32.87 & N/A & \underline{20.07} & 38.17 & \underline{32.23} & 37.74 & \textbf{21.56} & N/A \\
\bottomrule
\end{tabular}
}
\vspace{-2 mm}
\caption{Comparison with external systems: mean WER per test suite ($\downarrow$), model size, and inference cost. Parakeet is the teacher. Systems are grouped by size. Within each of the two middle groups, bold marks the best result per row and underline the second best. CAFE-8lang (last column) uses the same method, trained on the 8 MLS languages only, and is excluded from the group comparison. Per-dataset results are in Appendix~\ref{app:extfull}, Table~\ref{tab:fullref}.}
\label{tab:extsummary}
\vspace{-3 mm}
\end{table*}

\paragraph{Comparison with external systems.} Table~\ref{tab:extsummary} compares CAFE with strong public systems on all 48 test sets,\footnote{Open ASR Leaderboard English short-form (7) and long-form (4), FLEURS (5) and MLS (4), plus Multilingual LibriSpeech (8) and Common Voice v25 (24): 52 subsets, or 48 after removing the 4 MLS languages that appear twice.} with per-dataset WER and inference cost in Appendix~\ref{app:extfull} (Table~\ref{tab:fullref}). CAFE is small and fast, with 254M parameters, an RTFx of 213, and 2.6\,GB peak memory. Among compact systems it is the most accurate on the multilingual leaderboard sets, MLS-8, and Common Voice v25\footnote{Qwen3-ASR-0.6B has no 24-language Common Voice mean because some languages are outside its supported set.}, while OWSM v4 small is slightly better on English. On MLS-8 its WER of 6.58\% is lower than that of every external system, including Whisper-medium at 7.76\% and OWSM-CTC v4~\citep{peng2024owsmctc,peng2025owsm}. Larger models win on a few suites: Whisper-medium on FLEURS, and Qwen3-ASR-0.6B~\citep{shi2026qwen3} and OWSM-CTC v4 on English short-form. These models are three to four times larger than CAFE, and OWSM-CTC v4 uses about ten times its peak memory. CAFE thus lies on the Pareto front of accuracy and model size (Figure~\ref{fig:pareto}). The goal is not to rank systems, as external models tackle much broader scopes (e.g., omniASR~\cite{omnilingualasr2025} supports over 1,600 languages, while Qwen3-ASR offers strong Chinese dialect support). Instead, we aim to show that our student model and the teacher remain competitive for their parameter scale. This keeps our findings close to current state-of-the-art systems and to practical use.

The gap to the teacher depends on the test suite. CAFE matches the teacher on MLS-8, helped by checkpoint averaging, but trails it on English short-form, at 9.03\% versus 6.83\%, partly because the student sees less English data. We intentionally leave out corpora with potential legal concerns to ensure the model can be released (Appendix~\ref{app:data}).

\paragraph{Routing.} The adapters carry the multilingual gain and must be skipped for English. Applying them to every utterance raises EN-7 from 9.03\% to 15.95\%, and never applying them raises ML-9 from 7.00\% to 8.04\%. When the language is unknown, a small gate of 0.19\% parameters that we release makes this choice at a cost of 0.05 points in average WER (Appendix~\ref{app:gate}).

\begin{figure}[t]
    \centering
    \includegraphics[width=0.85\columnwidth]{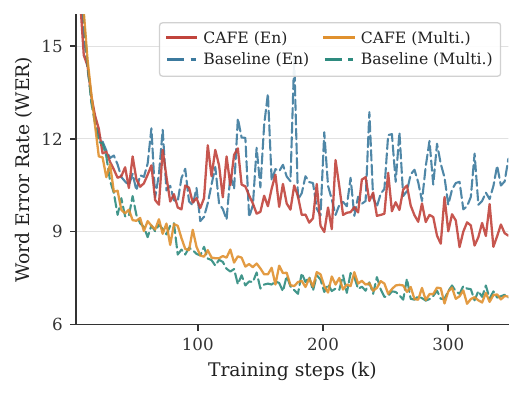}
    \vspace{-3 mm}
    \caption{English and multilingual test-set WER during training for Baseline-8lang and proposed CAFE-8lang.}
    \label{fig:curve}
    \vspace{-4 mm}
\end{figure}

\paragraph{Training dynamics.} Figure~\ref{fig:curve} tracks test-set WER during training for the 8-language pair. Multilingual WER falls smoothly for both models and the two curves stay close, so CAFE does not cost multilingual accuracy. English behaves differently. In the second half of training, the Baseline oscillates strongly, with spikes above 14\% WER, while CAFE stays lower, varies less, with a standard deviation of 0.62 versus 0.96, and keeps improving. These curves are decoded on the test sets for analysis only and are not used for model selection (Model selection in Section~\ref{sec:setup}).

\paragraph{Discussion.}\label{sec:analysis} The results agree with the analysis in Section~\ref{sec:analysis-motiv}. English bypasses the adapters and the shared encoder receives no CKA gradient, so English utterances contribute exactly the same losses in CAFE and in Capacity. The English gain over Capacity, from 9.67\% to 9.03\%, must therefore come through the shared parameters, which non-English data trains differently once CKA has shaped the adapters. The training curves support this: CAFE removes the late English drift seen in the Baseline. We therefore attribute the gain to less interference on the shared encoder late in training.

\section{Conclusion}

Feature distillation does not help uniformly. We used gradients to explain why feature distillation does not add to logit distillation in compact multilingual ASR. The projection-based gradient is orthogonal to the task. The CKA gradient is weakly aligned with it, lowering WER on non-English languages but raising it on English. Because this gradient overlaps almost entirely with the subspace English uses, masking the loss fails in training, and our diagnostics argue against gating or projecting it.
Guided by this diagnosis, we proposed Confined-Adapter Feature distillation (CAFE), which confines the CKA gradient to language-routed adapters that English bypasses and so moves the fix from the loss to the parameters. CAFE improves on the Baseline mainly in English, with a smaller multilingual gain. It also beats an adapter-only control and more than halves the variation of English WER across checkpoints.

\newpage

\section*{Limitations}

We initialize the student by structured pruning of the teacher. This is the standard way to build a compact student at this data scale, and prior work finds that it outperforms training from scratch \citep{zhang2025prunernnt,lin2026arkasr}. Our findings therefore apply to how such models are trained in practice, and we limit our claims to this setting. Whether the same pattern holds under random initialization is left to future work.

Each configuration is trained once. Our ablations already need several full runs on 564k hours of data, and repeating each with several seeds is too costly. As a check, the 8-language pair uses a different seed and language set and shows the same pattern (Section~\ref{sec:exp}). The significance tests in Appendix~\ref{app:sig} also compare the 7 checkpoints behind each averaged model, which captures some variation within a run. We acknowledge that these checks give only partial evidence and do not fully replace a multi-seed study.

CAFE needs to know whether an utterance is English. The main results take this from the test set, but in practice the language may be unknown. We therefore release a small language gate with the model (Appendix~\ref{app:gate}). It adds 0.19\% parameters and raises the average WER from 8.02\% to 8.07\%, still well below the 8.55\% of the Baseline.

\section*{Ethical Statement}
\paragraph{Data and privacy.} All training and evaluation corpora are public research datasets used under their licenses (listed in Appendix~\ref{app:data}). We collected no new data from human subjects and ran no annotation study. Speech can reveal speaker identity and other personal information, so we use only the released data, make no attempt to re-identify speakers, and follow each corpus's official train and test splits. For Common Voice we also remove training clips that share an utterance, speaker, or transcript with the test set (Appendix~\ref{app:data}).

\paragraph{Fairness and intended use.} ASR quality is uneven across languages, accents, and dialects, and compact models can widen this gap. Our study addresses this directly: it aims to help languages with less data than English without degrading English. We report dialectal English (CORAAL) and per-language results so that differences stay visible rather than hidden in an average. Gaps remain, and performance on languages, accents, or domains outside our test sets is not guaranteed. The intended use is transcription for accessibility and for on-device and offline settings. Like any ASR system it can be misused, for example for surveillance, and should be deployed with consent and appropriate safeguards.

\paragraph{Compute and efficiency.} Training at this scale (48 H100 GPUs, 564k hours of audio) has a significant one-time energy cost. The resulting student is compact (254M parameters), fast, and uses little memory, so the recurring cost of inference, which dominates in deployment, is low. Distilling a small model from a larger teacher is partly a step toward efficiency and accessibility.

\paragraph{Usage of AI assistants.} AI language models were used in two capacities: (1) code debugging, and (2) language polishing during paper writing. All experimental design, model development, analysis, and scientific conclusions were made by the authors.

\bibliography{custom}

\appendix

\section{Efficiency measurement protocol}
\label{app:efficiency}

We measure inference cost in four ways: model size (parameters), throughput (RTFx), latency per utterance, and peak GPU memory. All measurements use one GPU under fixed conditions, so that differences reflect the models rather than the hardware or I/O.

\begin{table*}[!t]
    \centering\fontsize{7.2pt}{8.2pt}\selectfont
    \setlength{\tabcolsep}{4pt}
    \renewcommand{\arraystretch}{0.62}
    \resizebox{\textwidth}{!}{%
    \begin{tabular}{@{}l r !{\vrule width 0.6pt} r r r r r !{\vrule width 0.6pt} r r r r !{\vrule width 0.6pt} r@{}}
    \toprule
     & \multicolumn{1}{c}{\makecell{Parakeet\\(teacher)}} & \multicolumn{1}{c}{\makecell{Whisper\\medium}} & \multicolumn{1}{c}{\makecell{omniASR-LLM\\300M v2}} & \multicolumn{1}{c}{\makecell{OWSM-\\CTC v4}} & \multicolumn{1}{c}{\makecell{Qwen3-ASR\\0.6B}} & \multicolumn{1}{c}{\makecell{omniASR-\\CTC-1B v2}} & \multicolumn{1}{c}{\makecell{Whisper\\small}} & \multicolumn{1}{c}{\makecell{omniASR-CTC\\300M v2}} & \multicolumn{1}{c}{\makecell{OWSM\\v4 small}} & \multicolumn{1}{c}{\textbf{\makecell{CAFE\\(ours)}}} & \multicolumn{1}{c}{\textbf{\makecell{CAFE-\\8lang}}} \\
    \midrule
    Number of languages & 25 & $\sim$100 & $>$1600 & 75 & 30 & $>$1600 & $\sim$100 & $>$1600 & 75 & 25 & 8 \\
    \midrule
    Params (M) $\downarrow$ & 627 & 764 & 1{,}632 & 1{,}011 & 782 & 976 & 242 & 326 & 370 & 254 & 254 \\
    RTFx bs1 $\uparrow$ & 198 & 21 & 10 & 590 & 18 & 513 & 35 & 848 & 39 & 213 & 213 \\
    RTFx bs32 $\uparrow$ & 2{,}520 & 386 & 117 & 1{,}282 & 310 & 2{,}142 & 546 & 3{,}405 & 37 & 2{,}602 & 2{,}602 \\
    P50 lat.\ bs1 (s) $\downarrow$ & 0.095 & 0.704 & 1.990 & 0.033 & 1.070 & 0.037 & 0.417 & 0.024 & 0.501 & 0.089 & 0.089 \\
    Peak mem (MB) $\downarrow$ & 3{,}310 & 6{,}932 & 17{,}310 & 25{,}803 & 6{,}266 & 16{,}033 & 2{,}617 & 14{,}802 & 2{,}505 & 2{,}593 & 2{,}593 \\
    \midrule\midrule
    \multicolumn{12}{l}{\emph{Open ASR Leaderboard - English short-form (7)}}\\
    ami & 11.44 & 38.96 & 26.03 & 13.12 & 10.96 & 24.24 & 53.35 & 32.51 & 15.47 & 13.91 & 12.99 \\
    earnings22 & 11.19 & 11.49 & 16.90 & 13.89 & 10.76 & 19.79 & 13.21 & 26.33 & 14.48 & 13.55 & 14.14 \\
    gigaspeech & 9.57 & 10.05 & 12.83 & 10.87 & 8.37 & 15.80 & 10.56 & 22.84 & 11.17 & 11.72 & 11.56 \\
    LibriSpeech.clean & 1.93 & 2.58 & 2.16 & 2.60 & 1.75 & 2.83 & 2.59 & 4.71 & 2.56 & 2.48 & 2.28 \\
    LibriSpeech.other & 3.59 & 4.87 & 6.54 & 4.88 & 4.08 & 6.67 & 6.60 & 11.86 & 5.98 & 5.81 & 5.19 \\
    spgispeech & 3.98 & 2.92 & 5.82 & 2.57 & 2.79 & 7.37 & 3.24 & 14.05 & 2.79 & 7.98 & 7.51 \\
    voxpopuli & 6.14 & 7.63 & 8.71 & 7.35 & 6.82 & 8.79 & 7.85 & 13.28 & 8.65 & 7.76 & 7.31 \\
    \rowcolor{avgcol} category mean (7) & 6.83 & 11.21 & 11.28 & \underline{7.90} & \textbf{6.50} & 12.21 & 13.91 & 17.94 & \textbf{8.73} & \underline{9.03} & 8.71 \\
    \midrule
    \multicolumn{12}{l}{\emph{Open ASR Leaderboard - English long-form (4 sets; CORAAL as 8 subsets)}}\\
    earnings21 & 10.21 & 12.52 & 21.30 & 9.82 & 9.20 & 52.02 & 14.07 & 31.14 & 10.80 & 12.31 & 11.99 \\
    earnings22-lf & 14.52 & 16.73 & 24.31 & 13.62 & 13.59 & 53.60 & 17.98 & 40.19 & 15.24 & 14.76 & 14.41 \\
    tedlium & 2.74 & 3.10 & 11.20 & 3.38 & 2.89 & 52.95 & 3.52 & 31.48 & 4.06 & 4.08 & 3.72 \\
    coraal avg (8) & 15.59 & 32.76 & 37.20 & 22.06 & 17.98 & 60.71 & 31.27 & 66.97 & 21.94 & 22.25 & 21.84 \\
    \quad ATL & 16.69 & 39.45 & 40.75 & 24.11 & 18.20 & 55.15 & 34.57 & 72.58 & 23.63 & 22.77 & 22.08 \\
    \quad DCA & 19.00 & 39.57 & 40.81 & 26.22 & 21.71 & 64.03 & 39.34 & 67.28 & 26.05 & 25.86 & 25.48 \\
    \quad DCB & 11.72 & 23.40 & 34.54 & 16.93 & 13.70 & 60.98 & 25.31 & 63.50 & 17.74 & 18.35 & 18.05 \\
    \quad DTA & 19.61 & 36.16 & 42.01 & 29.29 & 22.76 & 62.19 & 38.86 & 75.00 & 27.95 & 28.13 & 27.64 \\
    \quad LES & 15.47 & 33.06 & 36.95 & 21.43 & 18.04 & 58.62 & 29.94 & 65.49 & 20.86 & 21.77 & 21.96 \\
    \quad PRV & 22.36 & 43.39 & 46.82 & 31.41 & 26.92 & 72.71 & 41.13 & 79.68 & 30.64 & 31.01 & 30.36 \\
    \quad ROC & 8.16 & 19.41 & 23.31 & 10.96 & 9.41 & 64.91 & 16.46 & 54.70 & 11.61 & 12.75 & 12.36 \\
    \quad VLD & 11.71 & 27.64 & 32.38 & 16.11 & 13.11 & 47.10 & 24.51 & 57.57 & 17.06 & 17.38 & 16.82 \\
    \rowcolor{avgcol} category mean (4) & 10.77 & 16.28 & 23.50 & \underline{12.22} & \textbf{10.91} & 54.82 & 16.71 & 42.45 & \textbf{13.01} & \underline{13.35} & 12.99 \\
    \midrule
    \multicolumn{12}{l}{\emph{Open ASR Leaderboard - Multilingual: FLEURS (5)}}\\
    de & 4.25 & 4.66 & 7.99 & 8.26 & 5.54 & 7.46 & 7.61 & 14.53 & 10.08 & 7.61 & 6.79 \\
    es & 3.27 & 3.12 & 5.39 & 5.59 & 4.47 & 5.33 & 4.67 & 9.56 & 8.21 & 5.38 & 4.92 \\
    fr & 4.72 & 6.89 & 11.25 & 9.79 & 7.07 & 10.84 & 11.52 & 21.23 & 13.16 & 9.83 & 9.08 \\
    it & 2.45 & 3.56 & 5.47 & 5.76 & 4.62 & 4.87 & 16.02 & 9.46 & 6.87 & 4.43 & 4.05 \\
    pt & 4.54 & 4.29 & 7.70 & 7.42 & 6.08 & 7.19 & 6.27 & 13.63 & 8.76 & 7.74 & 7.14 \\
    \rowcolor{avgcol} category mean (5) & 3.85 & \textbf{4.50} & 7.56 & 7.36 & \underline{5.56} & 7.14 & \underline{9.22} & 13.68 & 9.42 & \textbf{7.00} & 6.40 \\
    \midrule
    \multicolumn{12}{l}{\emph{Open ASR Leaderboard - Multilingual: MLS (4)}}\\
    es & 4.40 & 4.77 & 4.80 & 6.44 & 6.80 & 4.20 & 6.89 & 7.25 & 7.61 & 4.74 & 4.32 \\
    fr & 5.13 & 7.17 & 6.82 & 8.31 & 8.54 & 6.94 & 10.93 & 12.95 & 11.05 & 4.85 & 4.31 \\
    it & 8.52 & 10.58 & 7.25 & 14.14 & 14.76 & 6.51 & 16.02 & 10.73 & 14.49 & 7.37 & 6.67 \\
    pt & 7.67 & 10.33 & 11.50 & 11.55 & 14.03 & 11.05 & 15.37 & 15.81 & 14.83 & 11.07 & 11.39 \\
    \rowcolor{avgcol} category mean (4) & 6.43 & 8.21 & \underline{7.59} & 10.11 & 11.03 & \textbf{7.18} & 12.30 & \underline{11.69} & 12.00 & \textbf{7.01} & 6.67 \\
    \midrule
    \rowcolor{avg2col} EN short-form \& Multilingual mean & 5.99 & 8.79 & 9.43 & \underline{8.32} & \textbf{7.40} & 9.68 & 12.34 & 15.31 & \underline{9.72} & \textbf{8.02} & 7.62 \\
    \midrule\midrule
    \multicolumn{12}{l}{\emph{Multilingual LibriSpeech (MLS-8)}}\\
    en & 4.82 & 6.21 & 8.68 & 6.75 & 5.77 & 9.20 & 7.66 & 15.10 & 7.96 & 6.48 & 6.05 \\
    de & 5.79 & 5.75 & 7.02 & 10.36 & 8.18 & 6.69 & 9.35 & 11.02 & 10.43 & 6.59 & 6.27 \\
    nl & 11.24 & 11.42 & 6.92 & 11.22 & 16.69 & 7.93 & 16.54 & 12.89 & 11.64 & 5.89 & 5.81 \\
    pl & 5.74 & 5.89 & 9.32 & 16.39 & 26.02 & 9.18 & 9.26 & 9.39 & 16.72 & 5.63 & 5.10 \\
    es & 4.39 & 4.76 & 4.80 & 6.44 & 6.80 & 4.20 & 6.90 & 7.25 & 7.61 & 4.74 & 4.33 \\
    fr & 5.07 & 7.18 & 6.82 & 8.31 & 8.54 & 6.94 & 10.94 & 12.95 & 11.05 & 4.87 & 4.32 \\
    it & 8.53 & 10.60 & 7.25 & 14.14 & 14.76 & 6.51 & 16.06 & 10.73 & 14.49 & 7.37 & 6.69 \\
    pt & 7.69 & 10.28 & 11.50 & 11.55 & 14.03 & 11.05 & 15.37 & 15.81 & 14.83 & 11.04 & 11.39 \\
    \rowcolor{avg2col} mean (8) & 6.66 & \underline{7.76} & 7.79 & 10.65 & 12.60 & \textbf{7.71} & \underline{11.51} & 11.89 & 11.84 & \textbf{6.58} & 6.25 \\
    \midrule\midrule
    \multicolumn{12}{l}{\emph{Common Voice v25 (24)}}\\
    bg & 18.21 & 23.76 & 12.87 & 46.54 & N/A & 16.64 & 41.00 & 30.08 & 52.54 & 18.20 & N/A \\
    cs & 5.92 & 18.30 & 9.61 & 23.05 & 42.79 & 10.53 & 33.55 & 20.33 & 27.39 & 12.47 & N/A \\
    da & 40.38 & 44.38 & 40.56 & 66.55 & 55.37 & 42.77 & 52.21 & 52.46 & 74.82 & 43.70 & N/A \\
    de & 4.40 & 7.83 & 10.58 & 8.83 & 8.30 & 10.02 & 13.97 & 19.29 & 10.71 & 8.83 & 7.78 \\
    el & 18.23 & 22.76 & 13.51 & 62.36 & 52.11 & 25.40 & 29.88 & 37.23 & 65.56 & 32.80 & N/A \\
    en & 8.47 & 12.38 & 18.94 & 11.45 & 9.67 & 19.78 & 17.18 & 31.20 & 14.22 & 13.94 & 12.95 \\
    es & 3.76 & 7.15 & 7.53 & 7.20 & 6.58 & 7.83 & 11.63 & 14.23 & 9.34 & 7.22 & 6.31 \\
    et & 22.29 & 44.33 & 22.62 & 53.75 & N/A & 20.79 & 69.50 & 39.42 & 52.14 & 27.37 & N/A \\
    fi & 48.76 & 53.14 & 51.97 & 64.14 & 72.67 & 52.67 & 58.76 & 58.49 & 67.07 & 54.29 & N/A \\
    fr & 7.33 & 18.85 & 14.28 & 12.31 & 11.75 & 15.18 & 26.92 & 24.82 & 16.22 & 11.64 & 10.46 \\
    hu & 6.55 & 18.90 & 11.02 & 35.12 & 55.36 & 13.44 & 36.10 & 26.10 & 36.78 & 11.79 & N/A \\
    it & 3.59 & 8.62 & 8.97 & 8.18 & 8.70 & 8.78 & 16.81 & 16.28 & 10.24 & 7.15 & 6.29 \\
    lt & 16.60 & 50.51 & 22.56 & 60.22 & N/A & 26.23 & 73.12 & 45.25 & 65.21 & 25.68 & N/A \\
    lv & 36.08 & 33.94 & 18.87 & 55.13 & N/A & 21.05 & 55.98 & 36.61 & 58.51 & 17.79 & N/A \\
    mt & 12.16 & 93.16 & 11.79 & 62.61 & N/A & 17.77 & 109.40 & 31.01 & 73.66 & 46.73 & N/A \\
    nl & 4.21 & 6.53 & 6.98 & 8.07 & 10.11 & 7.70 & 12.66 & 16.74 & 12.11 & 6.04 & 5.11 \\
    pl & 8.53 & 15.31 & 15.81 & 20.25 & 30.20 & 15.53 & 20.74 & 25.53 & 24.85 & 15.91 & 14.26 \\
    pt & 8.04 & 9.85 & 12.16 & 8.91 & 10.92 & 12.87 & 22.75 & 23.69 & 13.01 & 10.94 & 9.58 \\
    ro & 10.79 & 22.13 & 13.35 & 20.10 & 42.23 & 12.97 & 31.82 & 23.80 & 34.09 & 22.46 & N/A \\
    ru & 5.54 & 9.44 & 12.93 & 9.06 & 12.96 & 14.86 & 16.42 & 32.45 & 14.68 & 14.62 & N/A \\
    sk & 8.36 & 40.31 & 15.70 & 33.79 & N/A & 25.43 & 52.97 & 34.31 & 41.75 & 21.68 & N/A \\
    sl & 49.60 & 33.42 & 17.71 & 34.86 & N/A & 25.05 & 45.02 & 48.08 & 42.90 & 26.51 & N/A \\
    sv-SE & 17.59 & 12.13 & 17.90 & 35.41 & 37.47 & 17.22 & 21.15 & 33.36 & 41.65 & 17.99 & N/A \\
    uk & 30.09 & 39.44 & 37.12 & 41.03 & N/A & 41.06 & 46.62 & 52.84 & 46.21 & 41.70 & N/A \\
    \rowcolor{avg2col} mean all (24) & 16.48 & 26.94 & \textbf{17.72} & 32.87 & N/A & \underline{20.07} & 38.17 & \underline{32.23} & 37.74 & \textbf{21.56} & N/A \\
    \bottomrule
    \end{tabular}%
    }
    \caption{Comparison with external systems on all test sets: per-dataset WER and inference efficiency (Appendix~\ref{app:efficiency}). Parakeet is the teacher (reference). Systems are grouped by size, and bold and underline mark the best and second-best mean per row within the large group (Whisper-medium, omniASR-LLM-300M v2, OWSM-CTC v4, Qwen3-ASR-0.6B, omniASR-CTC-1B v2) and within the compact group (Whisper-small, omniASR-CTC-300M v2, OWSM v4 small, and CAFE). CAFE-8lang (last column) uses the same method, trained on the 8 MLS languages only to replicate the result at a smaller data scale; it is not included in the group comparison, has the same inference cost as CAFE, and is N/A for languages outside its set. Long-form results for omniASR and Qwen3-ASR-0.6B use 40\,s non-overlapping windows (the maximum input length of omniASR). Qwen3-ASR-0.6B was also run with its native 1200\,s windows, but degenerated (repetition loops or early stopping) with WER of roughly 90--115\%, so it is reported with the same 40\,s windows as omniASR.}
    \label{tab:fullref}
\end{table*}

\paragraph{Hardware.} All timings are on a single NVIDIA H100 (80\,GB). To account for run-to-run variation, we run the full measurement on three separate GPUs and report the mean.

\paragraph{Data.} We time on MLS-8. For each of the eight languages we draw 500 random test utterances (4000 in total), keeping only clips of at most 30 seconds, so that every model, including Whisper with its fixed 30-second window, runs on the same set. All clips are decoded to WAV in advance, so the timing includes only model computation. The same set is used for every model and GPU.

\paragraph{Metrics.} RTFx is the total audio duration divided by the total computation time (higher is better), reported at batch sizes 1 and 32. P50 latency is the median time per utterance at batch size 1 (lower is better). Peak memory is the maximum GPU memory used at batch size 32 (lower is better). All models run in the same code on the same data and batch sizes, so the numbers in Tables~\ref{tab:efficiency} and~\ref{tab:fullref} are directly comparable.

\section{Per-dataset results for external systems}
\label{app:extfull}

Table~\ref{tab:fullref} gives the per-dataset results behind Table~\ref{tab:extsummary}. As in the main text, the goal is not to rank systems, since several external models cover far more languages (over 1600 for omniASR), which is a harder task. The goal is to show that our student and the teacher are competitive for their size. All systems, including the teacher, are decoded and scored with the same code and the official Open ASR Leaderboard text normalizer, applied to both hypotheses and references. The teacher's SPGISpeech WER of 3.98\% is close to that of external systems such as Whisper-medium at 2.92\%, so a normalization mismatch does not explain the student's higher SPGISpeech WER. On the English short-form sets, the gap between teacher and student varies by domain, from 0.55 points on LibriSpeech clean to 4.0 points on SPGISpeech, which points to training data coverage rather than model size (Appendix~\ref{app:data}).

The four shared MLS languages use the same audio in both groups. For the teacher, Whisper, and our models, the two groups come from separate batched bf16 runs, so padding changes a few greedy outputs. Systems evaluated later decode each shared set once. We report the results as measured rather than replacing them. The two runs differ by 0.02 WER points on average and at most 0.06, which changes no comparison in this paper.

\begin{table}[t]
    \centering\small
    \setlength{\tabcolsep}{5pt}
    \begin{tabular}{l r r r}
    \toprule
    Lang. & \makebox[18mm][r]{Utterances} & \makebox[15mm][r]{Hours} & \makebox[\pbw][r]{\%} \\
    \midrule
    en & \makebox[18mm][r]{62,165,596} & \makebox[15mm][r]{200,020.7} & \pbar{1.000}{35.46\%} \\
    de & \makebox[18mm][r]{4,702,308} & \makebox[15mm][r]{26,909.2} & \pbar{0.178}{4.77\%} \\
    fr & \makebox[18mm][r]{4,511,551} & \makebox[15mm][r]{25,603.1} & \pbar{0.171}{4.54\%} \\
    es & \makebox[18mm][r]{4,354,218} & \makebox[15mm][r]{23,614.3} & \pbar{0.162}{4.19\%} \\
    it & \makebox[18mm][r]{3,183,254} & \makebox[15mm][r]{22,482.3} & \pbar{0.157}{3.99\%} \\
    pl & \makebox[18mm][r]{2,774,073} & \makebox[15mm][r]{20,770.6} & \pbar{0.148}{3.68\%} \\
    nl & \makebox[18mm][r]{2,986,014} & \makebox[15mm][r]{20,556.0} & \pbar{0.147}{3.64\%} \\
    cs & \makebox[18mm][r]{2,378,100} & \makebox[15mm][r]{18,180.1} & \pbar{0.136}{3.22\%} \\
    pt & \makebox[18mm][r]{3,168,809} & \makebox[15mm][r]{17,893.3} & \pbar{0.135}{3.17\%} \\
    ro & \makebox[18mm][r]{2,311,316} & \makebox[15mm][r]{17,736.6} & \pbar{0.134}{3.14\%} \\
    hu & \makebox[18mm][r]{2,357,560} & \makebox[15mm][r]{17,601.1} & \pbar{0.133}{3.12\%} \\
    el & \makebox[18mm][r]{2,315,274} & \makebox[15mm][r]{17,577.7} & \pbar{0.133}{3.12\%} \\
    bg & \makebox[18mm][r]{2,244,380} & \makebox[15mm][r]{17,109.3} & \pbar{0.131}{3.03\%} \\
    sv & \makebox[18mm][r]{2,087,355} & \makebox[15mm][r]{15,774.3} & \pbar{0.125}{2.80\%} \\
    fi & \makebox[18mm][r]{1,811,718} & \makebox[15mm][r]{13,907.4} & \pbar{0.116}{2.47\%} \\
    da & \makebox[18mm][r]{1,770,697} & \makebox[15mm][r]{13,364.3} & \pbar{0.113}{2.37\%} \\
    lt & \makebox[18mm][r]{1,769,457} & \makebox[15mm][r]{13,352.2} & \pbar{0.113}{2.37\%} \\
    lv & \makebox[18mm][r]{1,730,360} & \makebox[15mm][r]{12,468.6} & \pbar{0.109}{2.21\%} \\
    sk & \makebox[18mm][r]{1,501,791} & \makebox[15mm][r]{11,429.1} & \pbar{0.104}{2.03\%} \\
    sl & \makebox[18mm][r]{1,443,654} & \makebox[15mm][r]{10,990.3} & \pbar{0.102}{1.95\%} \\
    et & \makebox[18mm][r]{1,300,809} & \makebox[15mm][r]{9,822.3} & \pbar{0.096}{1.74\%} \\
    hr & \makebox[18mm][r]{1,004,671} & \makebox[15mm][r]{7,817.3} & \pbar{0.087}{1.39\%} \\
    mt & \makebox[18mm][r]{923,711} & \makebox[15mm][r]{7,118.3} & \pbar{0.083}{1.26\%} \\
    ru & \makebox[18mm][r]{2,008,596} & \makebox[15mm][r]{1,925.5} & \pbar{0.059}{0.34\%} \\
    uk & \makebox[18mm][r]{63,311} & \makebox[15mm][r]{77.7} & \pbar{0.050}{0.01\%} \\
    \midrule
    \rowcolor{avg2col} Total (25) & \makebox[18mm][r]{116,868,583} & \makebox[15mm][r]{564,101.6} & \makebox[\pbw][r]{100.0\%} \\
    \bottomrule
    \end{tabular}
    \caption{Training data per language, sorted by hours. Blue bars show each language's share of the total.}
    \label{tab:trainlang}
    \end{table}

\begin{table*}[!t]
\centering\small
\setlength{\tabcolsep}{4pt}
\renewcommand{\arraystretch}{1.05}
\begin{tabular}[t]{@{}l@{\hspace{6mm}}l@{}}
\begin{tabular}[t]{@{}llll@{}}
\toprule
\lc{Lang.} & \uc{Samples} & \hc{Hours} & \makebox[\pbw][r]{\%} \\
\midrule
\multicolumn{4}{@{}l}{\cellcolor{avgcol}\textbf{VoxPopuli}~{\footnotesize\emph{(23 langs, 373,686\,h, 66.24\%)}}} \\
\lc{\footnotesize en} & \uc{3,053,532} & \hc{24,103.2} & \pbar{1.000}{6.45\%} \\
\lc{\footnotesize de} & \uc{2,979,597} & \hc{23,198.5} & \pbar{0.950}{6.21\%} \\
\lc{\footnotesize fr} & \uc{2,955,798} & \hc{22,777.4} & \pbar{0.927}{6.10\%} \\
\lc{\footnotesize it} & \uc{2,814,827} & \hc{21,788.2} & \pbar{0.871}{5.83\%} \\
\lc{\footnotesize es} & \uc{2,788,598} & \hc{21,348.5} & \pbar{0.845}{5.71\%} \\
\lc{\footnotesize pl} & \uc{2,652,542} & \hc{20,556.1} & \pbar{0.802}{5.50\%} \\
\lc{\footnotesize nl} & \uc{2,474,233} & \hc{18,882.2} & \pbar{0.708}{5.05\%} \\
\lc{\footnotesize cs} & \uc{2,333,712} & \hc{18,125.7} & \pbar{0.666}{4.85\%} \\
\lc{\footnotesize ro} & \uc{2,296,650} & \hc{17,717.2} & \pbar{0.643}{4.74\%} \\
\lc{\footnotesize el} & \uc{2,303,443} & \hc{17,563.8} & \pbar{0.635}{4.70\%} \\
\lc{\footnotesize hu} & \uc{2,278,164} & \hc{17,484.2} & \pbar{0.630}{4.68\%} \\
\lc{\footnotesize pt} & \uc{2,282,071} & \hc{17,334.2} & \pbar{0.622}{4.64\%} \\
\lc{\footnotesize bg} & \uc{2,234,897} & \hc{17,095.7} & \pbar{0.607}{4.57\%} \\
\lc{\footnotesize sv} & \uc{2,054,815} & \hc{15,741.4} & \pbar{0.532}{4.21\%} \\
\lc{\footnotesize fi} & \uc{1,799,765} & \hc{13,894.5} & \pbar{0.430}{3.72\%} \\
\lc{\footnotesize da} & \uc{1,763,822} & \hc{13,356.2} & \pbar{0.399}{3.57\%} \\
\lc{\footnotesize lt} & \uc{1,754,809} & \hc{13,331.6} & \pbar{0.399}{3.57\%} \\
\lc{\footnotesize lv} & \uc{1,595,156} & \hc{12,292.1} & \pbar{0.340}{3.29\%} \\
\lc{\footnotesize sk} & \uc{1,476,106} & \hc{11,398.3} & \pbar{0.290}{3.05\%} \\
\lc{\footnotesize sl} & \uc{1,433,486} & \hc{10,979.5} & \pbar{0.267}{2.94\%} \\
\lc{\footnotesize et} & \uc{1,281,814} & \hc{9,788.0} & \pbar{0.200}{2.62\%} \\
\lc{\footnotesize hr} & \uc{1,004,410} & \hc{7,817.1} & \pbar{0.090}{2.09\%} \\
\lc{\footnotesize mt} & \uc{919,343} & \hc{7,112.8} & \pbar{0.050}{1.90\%} \\
\addlinespace[0.4em]
\multicolumn{4}{@{}l}{\cellcolor{avgcol}\textbf{YODAS}~{\footnotesize\emph{(23 langs, 104,306\,h, 18.49\%)}}} \\
\lc{\footnotesize en} & \uc{39,801,853} & \hc{100,102.2} & \pbar{1.000}{95.97\%} \\
\lc{\footnotesize ru} & \uc{1,873,248} & \hc{1,732.4} & \pbar{0.066}{1.66\%} \\
\lc{\footnotesize es} & \uc{947,615} & \hc{779.4} & \pbar{0.057}{0.75\%} \\
\lc{\footnotesize fr} & \uc{557,519} & \hc{715.5} & \pbar{0.057}{0.69\%} \\
\lc{\footnotesize de} & \uc{415,050} & \hc{496.2} & \pbar{0.055}{0.48\%} \\
\lc{\footnotesize pt} & \uc{727,788} & \hc{262.1} & \pbar{0.052}{0.25\%} \\
\lc{\footnotesize it} & \uc{86,580} & \hc{120.2} & \pbar{0.051}{0.12\%} \\
\lc{\footnotesize nl} & \uc{71,477} & \hc{37.8} & \pbar{0.050}{0.04\%} \\
\lc{\footnotesize pl} & \uc{28,678} & \hc{21.1} & \pbar{0.050}{0.02\%} \\
\lc{\footnotesize uk} & \uc{9,479} & \hc{10.2} & \pbar{0.050}{0.01\%} \\
\lc{\footnotesize hu} & \uc{6,530} & \hc{8.5} & \pbar{0.050}{0.01\%} \\
\lc{\footnotesize ro} & \uc{2,303} & \hc{5.8} & \pbar{0.050}{0.01\%} \\
\lc{\footnotesize fi} & \uc{4,637} & \hc{3.8} & \pbar{0.050}{0.00\%} \\
\lc{\footnotesize cs} & \uc{4,185} & \hc{3.8} & \pbar{0.050}{0.00\%} \\
\lc{\footnotesize sv} & \uc{3,189} & \hc{1.9} & \pbar{0.050}{0.00\%} \\
\lc{\footnotesize bg} & \uc{1,533} & \hc{1.8} & \pbar{0.050}{0.00\%} \\
\lc{\footnotesize el} & \uc{514} & \hc{1.0} & \pbar{0.050}{0.00\%} \\
\lc{\footnotesize lt} & \uc{390} & \hc{0.6} & \pbar{0.050}{0.00\%} \\
\lc{\footnotesize da} & \uc{656} & \hc{0.6} & \pbar{0.050}{0.00\%} \\
\lc{\footnotesize et} & \uc{513} & \hc{0.5} & \pbar{0.050}{0.00\%} \\
\lc{\footnotesize sk} & \uc{287} & \hc{0.5} & \pbar{0.050}{0.00\%} \\
\lc{\footnotesize hr} & \uc{261} & \hc{0.2} & \pbar{0.050}{0.00\%} \\
\lc{\footnotesize lv} & \uc{75} & \hc{0.1} & \pbar{0.050}{0.00\%} \\
\bottomrule
\end{tabular}
&
\begin{tabular}[t]{@{}llll@{}}
\toprule
\lc{Lang.} & \uc{Samples} & \hc{Hours} & \makebox[\pbw][r]{\%} \\
\midrule
\multicolumn{4}{@{}l}{\cellcolor{avgcol}\textbf{MLS}~{\footnotesize\emph{(8 langs, 50,686\,h, 8.99\%)}}} \\
\lc{\footnotesize en} & \uc{10,808,037} & \hc{44,659.7} & \pbar{1.000}{88.11\%} \\
\lc{\footnotesize de} & \uc{469,940} & \hc{1,966.5} & \pbar{0.090}{3.88\%} \\
\lc{\footnotesize nl} & \uc{374,287} & \hc{1,554.2} & \pbar{0.081}{3.07\%} \\
\lc{\footnotesize fr} & \uc{258,213} & \hc{1,076.6} & \pbar{0.071}{2.12\%} \\
\lc{\footnotesize es} & \uc{220,701} & \hc{917.7} & \pbar{0.067}{1.81\%} \\
\lc{\footnotesize it} & \uc{59,623} & \hc{247.4} & \pbar{0.053}{0.49\%} \\
\lc{\footnotesize pt} & \uc{37,533} & \hc{161.0} & \pbar{0.051}{0.32\%} \\
\lc{\footnotesize pl} & \uc{25,043} & \hc{103.6} & \pbar{0.050}{0.20\%} \\
\addlinespace[0.4em]
\multicolumn{4}{@{}l}{\cellcolor{avgcol}\textbf{Common Voice}~{\footnotesize\emph{(24 langs, 6,492\,h, 1.15\%)}}} \\
\lc{\footnotesize en} & \uc{1,500,870} & \hc{2,225.7} & \pbar{1.000}{34.28\%} \\
\lc{\footnotesize de} & \uc{837,721} & \hc{1,248.0} & \pbar{0.582}{19.22\%} \\
\lc{\footnotesize fr} & \uc{740,021} & \hc{1,033.6} & \pbar{0.490}{15.92\%} \\
\lc{\footnotesize es} & \uc{397,304} & \hc{568.7} & \pbar{0.291}{8.76\%} \\
\lc{\footnotesize it} & \uc{222,224} & \hc{326.5} & \pbar{0.188}{5.03\%} \\
\lc{\footnotesize ru} & \uc{135,348} & \hc{193.1} & \pbar{0.130}{2.97\%} \\
\lc{\footnotesize lv} & \uc{135,129} & \hc{176.4} & \pbar{0.123}{2.72\%} \\
\lc{\footnotesize pt} & \uc{121,417} & \hc{136.0} & \pbar{0.106}{2.09\%} \\
\lc{\footnotesize hu} & \uc{72,866} & \hc{108.4} & \pbar{0.094}{1.67\%} \\
\lc{\footnotesize pl} & \uc{67,810} & \hc{89.8} & \pbar{0.086}{1.38\%} \\
\lc{\footnotesize nl} & \uc{66,017} & \hc{81.8} & \pbar{0.083}{1.26\%} \\
\lc{\footnotesize uk} & \uc{53,832} & \hc{67.5} & \pbar{0.077}{1.04\%} \\
\lc{\footnotesize cs} & \uc{40,203} & \hc{50.6} & \pbar{0.069}{0.78\%} \\
\lc{\footnotesize et} & \uc{18,482} & \hc{33.8} & \pbar{0.062}{0.52\%} \\
\lc{\footnotesize sv} & \uc{29,351} & \hc{31.0} & \pbar{0.061}{0.48\%} \\
\lc{\footnotesize sk} & \uc{25,398} & \hc{30.3} & \pbar{0.061}{0.47\%} \\
\lc{\footnotesize lt} & \uc{14,258} & \hc{20.0} & \pbar{0.056}{0.31\%} \\
\lc{\footnotesize ro} & \uc{12,363} & \hc{13.6} & \pbar{0.054}{0.21\%} \\
\lc{\footnotesize el} & \uc{11,317} & \hc{12.9} & \pbar{0.053}{0.20\%} \\
\lc{\footnotesize bg} & \uc{7,950} & \hc{11.8} & \pbar{0.053}{0.18\%} \\
\lc{\footnotesize sl} & \uc{10,168} & \hc{10.8} & \pbar{0.053}{0.17\%} \\
\lc{\footnotesize fi} & \uc{7,316} & \hc{9.1} & \pbar{0.052}{0.14\%} \\
\lc{\footnotesize da} & \uc{6,219} & \hc{7.5} & \pbar{0.051}{0.12\%} \\
\lc{\footnotesize mt} & \uc{4,368} & \hc{5.5} & \pbar{0.050}{0.08\%} \\
\addlinespace[0.8em]
\midrule
\midrule
\multicolumn{4}{@{}l}{\cellcolor{avgcol}\emph{Monolingual English corpora}} \\
\midrule
\multicolumn{4}{@{}l}{\textbf{Libri-Light}~{\footnotesize\emph{(en, 23,735\,h, 4.21\%)}}} \\
\multicolumn{4}{@{}l}{\textbf{NSC}~{\footnotesize\emph{(en, 3,098\,h, 0.55\%)}}} \\
\multicolumn{4}{@{}l}{\textbf{Europarl-ASR}~{\footnotesize\emph{(en, 1,016\,h, 0.18\%)}}} \\
\multicolumn{4}{@{}l}{\textbf{LibriSpeech}~{\footnotesize\emph{(en, 961\,h, 0.17\%)}}} \\
\multicolumn{4}{@{}l}{\textbf{AMI}~{\footnotesize\emph{(en, 77\,h, 0.01\%)}}} \\
\multicolumn{4}{@{}l}{\textbf{VCTK}~{\footnotesize\emph{(en, 41\,h, 0.01\%)}}} \\
\multicolumn{4}{@{}l}{\strut} \\
\multicolumn{4}{@{}l}{\strut} \\
\multicolumn{4}{@{}l}{\strut} \\
\multicolumn{4}{@{}l}{\strut} \\
\multicolumn{4}{@{}l}{\strut} \\
\bottomrule
\end{tabular}
\end{tabular}
\par\vspace{4pt}
\begin{tabular}{@{}>{\columncolor{avg2col}}p{139mm}@{}}
\textbf{Total:}\quad 116,868,583 utterances \quad 564,101.6 hours \quad 25 languages \quad 10 corpora \\
\bottomrule
\end{tabular}
\caption{Training data by corpus and language. Each header gives a corpus with its number of languages, total hours, and share of all training data; the rows below give utterances, hours, and share within the corpus (blue bars) for each language. Left: VoxPopuli and YODAS. Right: MLS and Common Voice, then the six English-only corpora (Libri-Light, NSC, Europarl-ASR, LibriSpeech, AMI, VCTK). The green row is the total.}
\label{tab:traincorplang}
\end{table*}

\section{Training and evaluation data}
\label{app:data}

This appendix describes the training data (Tables~\ref{tab:trainlang} and~\ref{tab:traincorplang}), the test sets (Table~\ref{tab:evalfull}), and how we keep test material out of training. Our training data largely follows the teacher's. Because we plan to release our model, and to avoid commercial-use and redistribution risks, we leave out YouTube-Commons\footnote{\url{https://huggingface.co/datasets/PleIAs/YouTube-Commons}} and Fisher~\citep{cieri2004fisher}, and we use the Common Voice release that was current at the time of this work (v25)\footnote{\url{https://mozilladatacollective.com/datasets}}. The monolingual English corpora are Libri-Light~\citep{kahn2020librilight}, the Singapore English National Speech Corpus~\citep{koh2019nsc}, Europarl-ASR~\citep{garcesdiazmunio21_interspeech}, LibriSpeech~\citep{panayotov2015librispeech}, AMI~\citep{carletta2006ami}, and VCTK\footnote{\url{https://datashare.ed.ac.uk/items/30e7453c-9ea8-48b4-8e18-f96d0dc62928}}. Utterance counts and durations are those of the audio the student actually saw (after voice activity trimming for YODAS and segmentation for Libri-Light), so they differ slightly from the nominal durations. Leaving out Fisher (conversational telephone speech) and YouTube-Commons (varied web speech) mainly affects spontaneous, accented English with many named entities, such as SPGISpeech, Earnings, and AMI. This is consistent with the larger gap between teacher and student on these sets (Section~\ref{sec:exp}). All our systems use the same training and test data, so their comparison is not affected.

\begin{table*}[t]
\centering\small
\setlength{\tabcolsep}{5pt}
\renewcommand{\arraystretch}{0.98}
\begin{tabular}{@{}lll@{\hspace{3em}}lll@{}}
\toprule
\makebox[22mm][l]{\hspace{0.9em}Test set / Language} & \makebox[18mm][r]{Samples} & \makebox[14mm][r]{Hours} & \makebox[22mm][l]{Test set} & \makebox[18mm][r]{Samples} & \makebox[14mm][r]{Hours} \\
\midrule
\multicolumn{3}{@{}l}{\textbf{From Open ASR Leaderboard}} & \multicolumn{3}{@{}l}{\textbf{From the official dataset's test set}} \\
\midrule
\multicolumn{3}{@{}l}{\emph{English short-form (7)}} & \multicolumn{3}{@{}l}{\emph{Multilingual LibriSpeech (8)}} \\
\makebox[22mm][l]{\hspace{0.9em}ami} & \makebox[18mm][r]{11,653} & \makebox[14mm][r]{8.54} & \makebox[22mm][l]{\hspace{0.9em}en} & \makebox[18mm][r]{3,769} & \makebox[14mm][r]{15.55} \\
\makebox[22mm][l]{\hspace{0.9em}earnings22} & \makebox[18mm][r]{2,737} & \makebox[14mm][r]{5.43} & \makebox[22mm][l]{\hspace{0.9em}de} & \makebox[18mm][r]{3,394} & \makebox[14mm][r]{14.29} \\
\makebox[22mm][l]{\hspace{0.9em}gigaspeech} & \makebox[18mm][r]{19,898} & \makebox[14mm][r]{35.35} & \makebox[22mm][l]{\hspace{0.9em}nl} & \makebox[18mm][r]{628} & \makebox[14mm][r]{2.51} \\
\makebox[22mm][l]{\hspace{0.9em}LibriSpeech.clean} & \makebox[18mm][r]{2,620} & \makebox[14mm][r]{5.40} & \makebox[22mm][l]{\hspace{0.9em}pl} & \makebox[18mm][r]{284} & \makebox[14mm][r]{1.17} \\
\makebox[22mm][l]{\hspace{0.9em}LibriSpeech.other} & \makebox[18mm][r]{2,939} & \makebox[14mm][r]{5.34} & \makebox[22mm][l]{\hspace{0.9em}es} & \makebox[18mm][r]{1,527} & \makebox[14mm][r]{6.33} \\
\makebox[22mm][l]{\hspace{0.9em}spgispeech} & \makebox[18mm][r]{39,341} & \makebox[14mm][r]{100.00} & \makebox[22mm][l]{\hspace{0.9em}fr} & \makebox[18mm][r]{1,449} & \makebox[14mm][r]{5.95} \\
\makebox[22mm][l]{\hspace{0.9em}voxpopuli} & \makebox[18mm][r]{1,842} & \makebox[14mm][r]{4.93} & \makebox[22mm][l]{\hspace{0.9em}it} & \makebox[18mm][r]{657} & \makebox[14mm][r]{2.68} \\
\cellcolor{avgcol}\makebox[22mm][l]{subtotal} & \cellcolor{avgcol}\makebox[18mm][r]{81,030} & \cellcolor{avgcol}\makebox[14mm][r]{165.00} & \makebox[22mm][l]{\hspace{0.9em}pt} & \makebox[18mm][r]{192} & \makebox[14mm][r]{0.79} \\
\multicolumn{3}{@{}l}{\emph{English long-form (4)}} & \cellcolor{avgcol}\makebox[22mm][l]{subtotal} & \cellcolor{avgcol}\makebox[18mm][r]{11,900} & \cellcolor{avgcol}\makebox[14mm][r]{49.27} \\
\makebox[22mm][l]{\hspace{0.9em}earnings21} & \makebox[18mm][r]{44} & \makebox[14mm][r]{39.26} & \multicolumn{3}{@{}l}{\emph{Common Voice v25 (24)}} \\
\makebox[22mm][l]{\hspace{0.9em}earnings22-lf} & \makebox[18mm][r]{125} & \makebox[14mm][r]{119.89} & \makebox[22mm][l]{\hspace{0.9em}bg} & \makebox[18mm][r]{3,398} & \makebox[14mm][r]{5.25} \\
\makebox[22mm][l]{\hspace{0.9em}tedlium} & \makebox[18mm][r]{11} & \makebox[14mm][r]{2.47} & \makebox[22mm][l]{\hspace{0.9em}cs} & \makebox[18mm][r]{9,463} & \makebox[14mm][r]{12.13} \\
\makebox[22mm][l]{\hspace{0.9em}coraal (8)} & \makebox[18mm][r]{271} & \makebox[14mm][r]{160.94} & \makebox[22mm][l]{\hspace{0.9em}da} & \makebox[18mm][r]{2,654} & \makebox[14mm][r]{3.41} \\
\makebox[22mm][l]{\hspace{1.8em}ATL} & \makebox[18mm][r]{14} & \makebox[14mm][r]{8.62} & \makebox[22mm][l]{\hspace{0.9em}de} & \makebox[18mm][r]{16,206} & \makebox[14mm][r]{27.96} \\
\makebox[22mm][l]{\hspace{1.8em}DCA} & \makebox[18mm][r]{74} & \makebox[14mm][r]{34.04} & \makebox[22mm][l]{\hspace{0.9em}el} & \makebox[18mm][r]{1,735} & \makebox[14mm][r]{2.10} \\
\makebox[22mm][l]{\hspace{1.8em}DCB} & \makebox[18mm][r]{63} & \makebox[14mm][r]{46.06} & \makebox[22mm][l]{\hspace{0.9em}en} & \makebox[18mm][r]{16,398} & \makebox[14mm][r]{27.17} \\
\makebox[22mm][l]{\hspace{1.8em}DTA} & \makebox[18mm][r]{40} & \makebox[14mm][r]{25.14} & \makebox[22mm][l]{\hspace{0.9em}es} & \makebox[18mm][r]{15,902} & \makebox[14mm][r]{26.98} \\
\makebox[22mm][l]{\hspace{1.8em}LES} & \makebox[18mm][r]{15} & \makebox[14mm][r]{8.44} & \makebox[22mm][l]{\hspace{0.9em}et} & \makebox[18mm][r]{2,890} & \makebox[14mm][r]{5.35} \\
\makebox[22mm][l]{\hspace{1.8em}PRV} & \makebox[18mm][r]{32} & \makebox[14mm][r]{13.96} & \makebox[22mm][l]{\hspace{0.9em}fi} & \makebox[18mm][r]{1,566} & \makebox[14mm][r]{2.22} \\
\makebox[22mm][l]{\hspace{1.8em}ROC} & \makebox[18mm][r]{19} & \makebox[14mm][r]{13.19} & \makebox[22mm][l]{\hspace{0.9em}fr} & \makebox[18mm][r]{16,201} & \makebox[14mm][r]{26.51} \\
\makebox[22mm][l]{\hspace{1.8em}VLD} & \makebox[18mm][r]{14} & \makebox[14mm][r]{11.48} & \makebox[22mm][l]{\hspace{0.9em}hu} & \makebox[18mm][r]{12,989} & \makebox[14mm][r]{20.33} \\
\cellcolor{avgcol}\makebox[22mm][l]{subtotal} & \cellcolor{avgcol}\makebox[18mm][r]{451} & \cellcolor{avgcol}\makebox[14mm][r]{322.56} & \makebox[22mm][l]{\hspace{0.9em}it} & \makebox[18mm][r]{15,184} & \makebox[14mm][r]{26.49} \\
\multicolumn{3}{@{}l}{\emph{Open ASR Leaderboard - Multilingual (9)}} & \makebox[22mm][l]{\hspace{0.9em}lt} & \makebox[18mm][r]{5,644} & \makebox[14mm][r]{8.18} \\
\multicolumn{3}{@{}l}{\hspace{0.9em}\emph{Multilingual LibriSpeech (4)}} & \makebox[22mm][l]{\hspace{0.9em}lv} & \makebox[18mm][r]{7,820} & \makebox[14mm][r]{12.50} \\
\makebox[22mm][l]{\hspace{1.8em}es} & \makebox[18mm][r]{1,527} & \makebox[14mm][r]{6.33} & \makebox[22mm][l]{\hspace{0.9em}mt} & \makebox[18mm][r]{1,669} & \makebox[14mm][r]{2.36} \\
\makebox[22mm][l]{\hspace{1.8em}fr} & \makebox[18mm][r]{1,449} & \makebox[14mm][r]{5.95} & \makebox[22mm][l]{\hspace{0.9em}nl} & \makebox[18mm][r]{12,247} & \makebox[14mm][r]{16.66} \\
\makebox[22mm][l]{\hspace{1.8em}it} & \makebox[18mm][r]{657} & \makebox[14mm][r]{2.68} & \makebox[22mm][l]{\hspace{0.9em}pl} & \makebox[18mm][r]{10,004} & \makebox[14mm][r]{14.29} \\
\makebox[22mm][l]{\hspace{1.8em}pt} & \makebox[18mm][r]{192} & \makebox[14mm][r]{0.79} & \makebox[22mm][l]{\hspace{0.9em}pt} & \makebox[18mm][r]{9,670} & \makebox[14mm][r]{12.95} \\
\multicolumn{3}{@{}l}{\hspace{0.9em}\emph{FLEURS (5)}} & \makebox[22mm][l]{\hspace{0.9em}ro} & \makebox[18mm][r]{3,946} & \makebox[14mm][r]{4.61} \\
\makebox[22mm][l]{\hspace{1.8em}de} & \makebox[18mm][r]{862} & \makebox[14mm][r]{3.15} & \makebox[22mm][l]{\hspace{0.9em}ru} & \makebox[18mm][r]{10,283} & \makebox[14mm][r]{16.03} \\
\makebox[22mm][l]{\hspace{1.8em}es} & \makebox[18mm][r]{908} & \makebox[14mm][r]{3.09} & \makebox[22mm][l]{\hspace{0.9em}sk} & \makebox[18mm][r]{5,965} & \makebox[14mm][r]{7.18} \\
\makebox[22mm][l]{\hspace{1.8em}fr} & \makebox[18mm][r]{676} & \makebox[14mm][r]{1.95} & \makebox[22mm][l]{\hspace{0.9em}sl} & \makebox[18mm][r]{1,341} & \makebox[14mm][r]{1.68} \\
\makebox[22mm][l]{\hspace{1.8em}it} & \makebox[18mm][r]{865} & \makebox[14mm][r]{3.52} & \makebox[22mm][l]{\hspace{0.9em}sv-SE} & \makebox[18mm][r]{5,505} & \makebox[14mm][r]{6.96} \\
\makebox[22mm][l]{\hspace{1.8em}pt} & \makebox[18mm][r]{919} & \makebox[14mm][r]{3.24} & \makebox[22mm][l]{\hspace{0.9em}uk} & \makebox[18mm][r]{10,133} & \makebox[14mm][r]{14.00} \\
\cellcolor{avgcol}\makebox[22mm][l]{subtotal} & \cellcolor{avgcol}\makebox[18mm][r]{8,055} & \cellcolor{avgcol}\makebox[14mm][r]{30.70} & \cellcolor{avgcol}\makebox[22mm][l]{subtotal} & \cellcolor{avgcol}\makebox[18mm][r]{198,813} & \cellcolor{avgcol}\makebox[14mm][r]{303.30} \\
\midrule
\rowcolor{avg2col}\multicolumn{6}{@{}l}{\textbf{Total (48 unique test sets):} 24 languages, 296{,}424 samples, 855.08 hours} \\
\rowcolor{avg2col}\multicolumn{6}{@{}l}{\footnotesize Excludes the 4 Multilingual LibriSpeech leaderboard sets duplicated in Multilingual LibriSpeech (8).} \\
\bottomrule
\end{tabular}
\vspace{-1 mm}
\caption{Per-subset evaluation statistics for all 48 unique test subsets (samples and audio hours), grouped by source; blue \emph{subtotal} rows give per-group totals. The four Open ASR Leaderboard Multilingual LibriSpeech languages reuse the same audio as the corresponding Multilingual LibriSpeech (8) languages.}
\label{tab:evalfull}
\vspace{-1 mm}
\end{table*}

\paragraph{Training languages.} Table~\ref{tab:trainlang} lists the 25 training languages by hours. English dominates (about 35\%, 200k hours), followed by European languages at roughly 2 to 5\% each and a tail of lower-resource languages. The imbalance between English and the other languages therefore comes from the data, not from our design.

\paragraph{Training corpora.} Table~\ref{tab:traincorplang} breaks the same data down by corpus. The left panel shows the two largest corpora, VoxPopuli and YODAS. The right panel shows MLS and Common Voice, followed by the English corpora. The total matches Table~\ref{tab:trainlang}.

\paragraph{Test sets.} We combine the public Open ASR Leaderboard test sets with the larger multilingual Common Voice v25 (24 languages) and Multilingual LibriSpeech (MLS-8) test sets. The English part draws on LibriSpeech~\citep{panayotov2015librispeech}, GigaSpeech~\citep{chen2021gigaspeech}, SPGISpeech~\citep{oneill2021spgispeech}, VoxPopuli~\citep{wang2021voxpopuli}, AMI~\citep{carletta2006ami}, TED-LIUM~\citep{hernandez2018tedlium}, Earnings-21 and Earnings-22~\citep{delrio2021earnings21,delrio2022earnings22}, and CORAAL~\citep{kendall2023coraal}; the multilingual part uses FLEURS~\citep{conneau2023fleurs}, MLS~\citep{pratap2020mls}, and Common Voice~\citep{ardila2020commonvoice}. The four leaderboard MLS sets (es, fr, it, pt) use the same audio as the corresponding MLS-8 sets.

\paragraph{Test-set isolation.} No development or test data is used for training. For every corpus except Common Voice we follow the official splits. The official Common Voice train split is deliberately small, so we enlarge it with validated clips that the split leaves out. We then remove any clip that shares with a test example (i) its utterance ID, (ii) its speaker ID (client\_id), or (iii) its transcript after lowercasing and whitespace normalization. The speaker and transcript checks are needed because Common Voice records the same sentence from many speakers. Removing only matching clip IDs would still let test sentences and test speakers into training. This keeps a large amount of otherwise unused audio while removing any utterance, speaker, or transcript overlap with the test set.

\section{Robustness to the checkpoint-averaging window}
\label{app:window}

\begin{table}[h]
\centering\small
\begin{tabular}{ll ccc}
\toprule
system & $W$ & EN-7 & ML-9 & Avg \\
\midrule
\multirow{3}{*}{\textbf{CAFE}} & 5 & 9.10 & 7.02 & 8.06 \\
 & 7 & 9.03 & 7.00 & 8.02 \\
 & 10 & 9.05 & 7.03 & 8.04 \\
\midrule
\multirow{3}{*}{Baseline} & 5 & 9.85 & 7.25 & 8.55 \\
 & 7 & 9.85 & 7.24 & 8.55 \\
 & 10 & 9.99 & 7.17 & 8.58 \\
\bottomrule
\end{tabular}
\caption{Robustness to the CA window size: EN-7 / ML-9 / Avg for the last $W\in\{5,7,10\}$ checkpoints.}
\label{tab:nrobust}
\end{table}

The main results do not depend on the exact checkpoint-averaging (CA) window. Table~\ref{tab:nrobust} recomputes CAFE and the Baseline with the last 5, 7, and 10 checkpoints (saved every 6k steps). CAFE changes very little across windows, with an average WER between 8.02\% and 8.06\%, and stays ahead of the Baseline at every window. The Baseline's English WER rises from 9.85\% to 9.99\% with the widest window, in line with its larger variation across checkpoints, while multilingual WER is stable for both. The last-7 window used throughout is therefore representative, not tuned.

\section{Language gate for deployment}
\label{app:gate}

CAFE applies the adapters only to non-English utterances. The results in the main text take this decision from the language of each test set, but in deployment the language is often unknown. We therefore train a small gate that predicts it from the audio.

\begin{table}[t]
\centering\footnotesize
\begin{tabular}{@{}ccc@{}}
\toprule
Overall & English & Non-English \\
\midrule
98.9\% & 97.6\% & 99.9\% \\
\bottomrule
\end{tabular}
\caption{Gate accuracy on 7,692 utterances: 500 randomly sampled per test set from the Open ASR Leaderboard EN-7 and ML-9 test sets (all 192 for MLS pt, which has fewer than 500).}
\label{tab:gateacc}
\end{table}

\begin{table}[t]
\centering\footnotesize
\setlength{\tabcolsep}{3pt}
\resizebox{\columnwidth}{!}{%
\begin{tabular}{@{}l ccccc@{}}
\toprule
 & Baseline & \makecell{CAFE\\(w/ label)} & \makecell{CAFE + gate\\(w/o label)} & \makecell{Adapters\\always on} & \makecell{Adapters\\never on} \\
\midrule
EN-7 & 9.85 & 9.03 & 9.11 & 15.95 & 9.03 \\
\midrule
\ FLEURS (5) & 7.22 & 7.00 & 7.04 & 7.00 & 8.46 \\
\ MLS (4) & 7.27 & 7.01 & 7.01 & 7.01 & 7.62 \\
ML-9 & 7.24 & 7.00 & 7.03 & 7.00 & 8.04 \\
\rowcolor{avg2col} Avg (EN, ML) & 8.55 & 8.02 & 8.07 & 11.48 & 8.54 \\
\bottomrule
\end{tabular}%
}
\caption{WER (last-7 CA) of CAFE when the EN/non-EN decision is taken from the test set (w/ label) or made by the gate (w/o label), with the Baseline for reference. The last two columns apply the adapters to every utterance or to none.}
\label{tab:gate}
\end{table}

The gate reads the output of encoder layer 11 of the trained CAFE model, with the adapters off, and predicts whether an utterance is non-English. It uses attentive statistics pooling over time followed by a two-layer MLP, with 0.495M parameters in total, or 0.19\% of the model. The ASR model is frozen. Only the gate is trained, on the same training data, with labels from the language tag of each utterance. The adapters are applied when the predicted probability of non-English exceeds a threshold $\theta$. The gate is released together with the model checkpoints.\footnote{\url{https://huggingface.co/qvac/TranscribePsy-250M-EU}. The Usage section of the model card explains how to run the model with the gate. TranscribePsy-250M-EU refers to the 25-language CAFE model in this work.}

The gate reaches 98.9\% accuracy (Table~\ref{tab:gateacc}). Table~\ref{tab:gate} gives the resulting WER, scored as in Table~\ref{tab:full}. The last two columns show why a gate is needed. Applying the adapters to every utterance badly hurts English, raising EN-7 from 9.03\% to 15.95\%. Never applying them loses the multilingual gain, raising ML-9 from 7.00\% to 8.04\%, worse than the Baseline. 
A wrong activation on English costs more than a missed one on other languages, so we tune the threshold on held-out training-corpus samples not used to train the gate. We select $\theta{=}0.8582$ and use it for all test evaluations and the released model. With the gate, the average WER rises by only 0.05 points, from 8.02\% to 8.07\%. The gated model stays ahead of the Baseline on English, multilingual, and average WER.

\section{Significance testing}
\label{app:sig}

We test English, where the conflict we study appears, and the multilingual leaderboard sets. Each test compares two systems on the same utterances. At the utterance level we use a paired permutation test (10,000 draws), which swaps the outputs of the two systems on each utterance with probability 0.5, and a paired bootstrap (10,000 resamples within each test set) for the 95\% confidence interval. Both use the same means as Table~\ref{tab:full}: the mean of per-set WER for EN-7, and for ML-9 the mean of the FLEURS and MLS means. At the checkpoint level we compare the 7 checkpoints behind each averaged model, paired by order, with a sign test and a Wilcoxon signed-rank test.

\begin{table}[t]
\centering\footnotesize
\setlength{\tabcolsep}{3pt}
\resizebox{\columnwidth}{!}{%
\begin{tabular}{@{}l c c c c c@{}}
\toprule
EN-7 & $\Delta$WER & 95\% CI & \makecell{Perm.\\$p$} & Ckpt & \makecell{Wilcoxon\\$p$} \\
\midrule
CKA vs Baseline & $+1.40$ & $[+1.25,+1.56]$ & $10^{-4}$ & 0/7 & 0.016 \\
CAFE vs Baseline & $-0.82$ & $[-0.94,-0.71]$ & $10^{-4}$ & 7/7 & 0.016 \\
CAFE vs Capacity & $-0.64$ & $[-0.72,-0.56]$ & $10^{-4}$ & 6/7 & 0.047 \\
\midrule
ML-9 & & & & & \\
\midrule
CKA vs Baseline & $-0.42$ & $[-0.54,-0.29]$ & $10^{-4}$ & 7/7 & 0.016 \\
CAFE vs Baseline & $-0.24$ & $[-0.39,-0.08]$ & 0.0028 & 5/7 & 0.156 \\
CAFE vs Capacity & $-0.34$ & $[-0.47,-0.20]$ & $10^{-4}$ & 6/7 & 0.031 \\
\bottomrule
\end{tabular}%
}
\caption{Significance on EN-7 and ML-9. $\Delta$WER is the difference between the averaged models (negative means the first system is better), with its 95\% bootstrap CI and permutation $p$ ($10^{-4}$ is the smallest value 10,000 draws can give). Ckpt is the number of the 7 checkpoint pairs on which the first system is better; Wilcoxon is over the same pairs.}
\label{tab:sig}
\end{table}

The first two EN-7 rows of Table~\ref{tab:sig} are the two halves of our claim: CKA hurts English, and CAFE removes that harm. The third matters most. CAFE and Capacity have the same adapters and parameter count and differ only in the confined feature loss, so the gap comes from that loss rather than from extra capacity. With 7 checkpoints, the two-sided sign test cannot go below 0.016, and 6 of 7 gives 0.125; this reflects the small number of checkpoints. The Wilcoxon test, which also uses the size of each difference, gives 0.047.

On ML-9, CKA lowers WER at both levels, consistent with its multilingual gain in Section~\ref{sec:conditional}. CAFE is better than both the Baseline and Capacity at the utterance level. Against Capacity the checkpoint level agrees, with a Wilcoxon $p$ of 0.031. Overall, the tests agree with the main results: CAFE's gain comes mainly from English, and its multilingual gain is smaller. Since the 7 checkpoints come from late training, the consistent English gain also matches the pattern in Figure~\ref{fig:curve}, where English varies less for CAFE in the second half of training than for the Baseline.
\end{document}